\documentclass[10pt,twocolumn,letterpaper]{article}

\usepackage{cvpr}
\usepackage{times}
\usepackage{epsfig}
\usepackage{graphicx}
\usepackage{amsmath}
\usepackage{amssymb}

\usepackage{multirow}
\usepackage{graphicx}
\usepackage{subfig}
\usepackage{array}
\usepackage{tablefootnote}
\usepackage{threeparttable}
\usepackage{pifont}
\newcommand{\cmark}{\ding{51}}%
\newcommand{\xmark}{\ding{55}}%
\newcommand\rurl[1]{%
  \href{https://#1}{\nolinkurl{#1}}%
}
\usepackage[font=small]{caption}

\usepackage[pagebackref=true,breaklinks=true,letterpaper=true,colorlinks,bookmarks=false]{hyperref}

\cvprfinalcopy 

\def\cvprPaperID{136} 

\begin{document}

\title{LiteFlowNet: A Lightweight Convolutional Neural Network \\for Optical Flow Estimation}

\author{Tak-Wai Hui, Xiaoou Tang, Chen Change Loy\\
CUHK-SenseTime Joint Lab, The Chinese University of Hong Kong\\
{\tt\small \{twhui,xtang,ccloy\}@ie.cuhk.edu.hk}}

\maketitle

\begin{abstract}
FlowNet2~\cite{Ilg17}, the state-of-the-art convolutional neural network (CNN) for optical flow estimation, requires over 160M parameters to achieve accurate flow estimation. In this paper we present an alternative network that outperforms FlowNet2 on the challenging Sintel final pass and KITTI benchmarks, while being 30 times smaller in the model size and 1.36 times faster in the running speed. This is made possible by drilling down to architectural details that might have been missed in the current frameworks: (1) We present a more effective flow inference approach at each pyramid level through a lightweight cascaded network. It not only improves flow estimation accuracy through early correction, but also permits seamless incorporation of descriptor matching in our network. (2) We present a novel flow regularization layer to ameliorate the issue of outliers and vague flow boundaries by using a feature-driven local convolution. (3) Our network owns an effective structure for pyramidal feature extraction and embraces feature warping rather than image warping as practiced in FlowNet2. Our code and trained models are available at \url{https://github.com/twhui/LiteFlowNet}.
\end{abstract}

\section{Introduction}
\label{sec:introduction}
%
Optical flow estimation is a long-standing problem in computer vision. Due to the well-known aperture problem, optical flow is not directly measurable~\cite{Horn81, Hui13}. Hence, the estimation is typically solved by energy minimization in a coarse-to-fine framework~\cite{Brox04, Papenberg06, Brox11, Zimmer11, Sun14, Revaud15}. 
This class of techniques, however, involves complex energy optimization and thus it is not scalable for applications that demand real-time estimation.

FlowNet~\cite{Fischer15} and its successor FlowNet2~\cite{Ilg17}, have marked a milestone by using CNN for optical flow estimation. Their accuracies especially the successor are approaching that of state-of-the-art energy minimization approaches, while the speed is several orders of magnitude faster.
To push the envelop of accuracy, FlowNet2 is designed as a cascade of variants of FlowNet that each network in the cascade refines the preceding flow field by contributing on the flow increment between the first image and the warped second image. The model, as a result, comprises over 160M parameters, which could be formidable in many applications.
A recent network termed SPyNet~\cite{Ranjan17} attempts a network with smaller size of 1.2M parameters by adopting image warping in each pyramid level. Nonetheless, the accuracy can only match that of FlowNet but not FlowNet2. 
The objective of this study is to explore alternative CNN architectures for accurate flow estimation yet with high efficiency.
Our work is inspired by the successes of FlowNet2 and SPyNet, but we further drill down the key elements to fully unleash the potential of deep convolutional network combined with classical principles. 

\begin{figure}[t]
\centering
   \includegraphics[width=\linewidth]{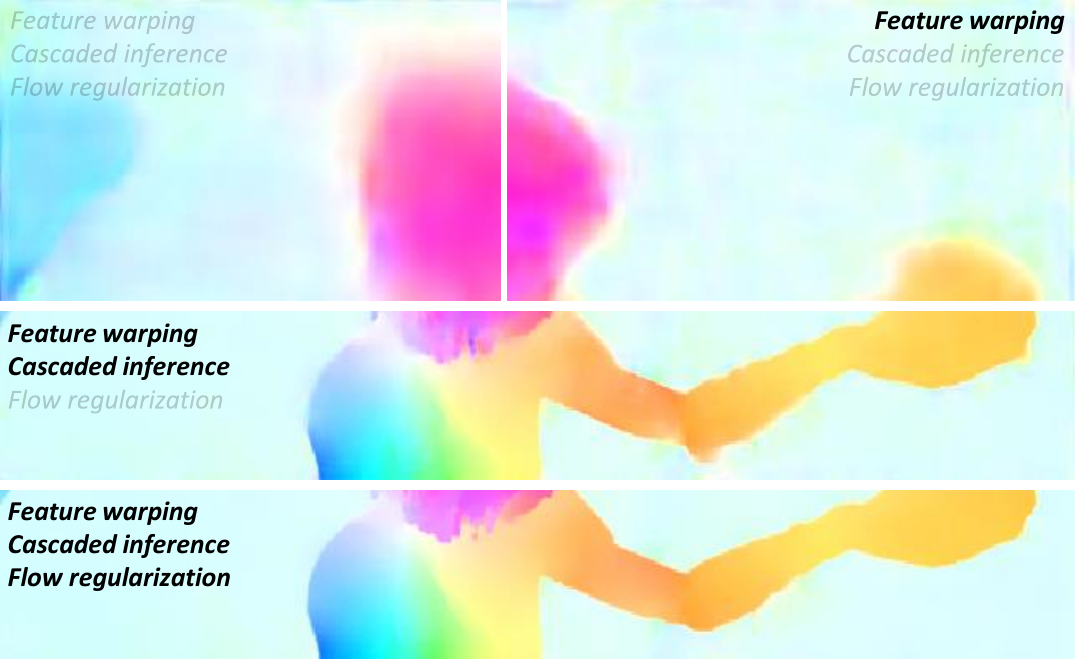}
\caption{Examples demonstrate the effectiveness of the proposed components in LiteFlowNet for i) feature warping, ii) cascaded flow inference, and iii) flow regularization. Enabled components are indicated with bold black fonts.}
\label{fig:overview}
\end{figure}

There are two general principles to improve the design of FlowNet2 and SPyNet. 
The first principle is \textit{pyramidal feature extraction}. The proposed network, dubbed \textit{LiteFlowNet}, consists of an encoder and a decoder. The encoder maps the given image pair, respectively, into two pyramids of multi-scale high-dimensional features. 
The decoder then estimates the flow field in a coarse-to-fine framework. At each pyramid level, the decoder infers the flow field by selecting and using the features of the same resolution from the feature pyramids.
This design leads to a lighter network compared to FlowNet2 that adopts U-Net architecture~\cite{Ronneberger15} for flow inference. In comparison to SPyNet, our network separates the process of feature extraction and flow estimation. This helps us to better pinpoint the bottleneck of accuracy and model size.

The second general principle is \textit{feature warping}.
FlowNet2 and SPyNet warp the second image towards the first image in the pair using the previous flow estimate, and then refine the estimate using the feature maps generated by the warped and the first images. Warping an image and then generating the feature maps of the warped image are two ordered steps. We find that the two steps can be reduced to a single one by directly warping the feature maps of the second image, which have been computed by the encoder. This one-step feature warping process reduces the more discriminative feature-space distance instead of the RGB-space distance between the two images. This makes our network more powerful and efficient in addressing the flow problem.

We now highlight the more specific differences between our network and existing CNN-based optical flow estimation frameworks: 

\noindent
1) \textit{Cascaded flow inference} -- 
At each pyramid level, we introduce a novel cascade of two lightweight networks. Each of them has a feature warping (f-warp) layer to displace the feature maps of the second image towards the first image using the flow estimate from the previous level. Flow residue is computed to further reduce the feature-space distance between the images.
This design is advantageous to the conventional design of using a single network for flow inference. 
First, the cascade progressively improves flow accuracy thus allowing an early correction of the estimate without passing more errors to the next level.
Second, this design allows seamless integration with descriptor matching. We assign a matching network to the first inference. Consequently, pixel-accuracy flow field can be generated first and then refined to sub-pixel accuracy in the subsequent inference network.
Since at each pyramid level the feature-space distance between the images has been reduced by feature warping, we can use a rather short displacement than~\cite{Fischer15, Ilg17} to establish the cost volume. Besides, matching is performed only at sampled positions and thus a sparse cost-volume is aggregated. This effectively reduces the computational burden raised by the explicit matching. 

\noindent
2) \textit{Flow regularization} -- 
The cascaded flow inference resembles the role of data fidelity in energy minimization methods. Using data term alone, vague flow boundaries and undesired artifacts exist in flow fields. To tackle this problem, local flow consistency and co-occurrence between flow boundaries and intensity edges are commonly used as the cues to regularize flow field. Some of the representative methods include anisotropic image-driven~\cite{Werlberger09}, image- and flow-driven~\cite{Sun08}, and complementary~\cite{Zimmer11} regularizations.
After cascaded flow inference, we allow the flow field to be further regularized by our novel feature-driven local convolution (f-lconv) layer\footnote{We name it as \textit{feature-driven local convolution} (f-lconv) layer in order to distinguish it from local convolution (lconv) layer of which filter weights are locally fixed in conventional CNNs~\cite{Taigman14}.} at each pyramid level.
The kernels of such a local convolution are adaptive to the pyramidal features from the encoder, flow estimate and occlusion probability map. This makes the flow regularization to be both flow- and image-aware.  
To our best knowledge, state-of-the-art CNNs do not explore such a flow regularization.

The effectiveness of the aforementioned contributions are depicted in Figure~\ref{fig:overview}. 
In summary, we propose a compact LiteFlowNet to estimate optical flow. Our network innovates the useful elements from conventional methods. \eg, brightness constraint in data fidelity to pyramidal CNN features and image warping to CNN feature warping. More specifically, we present a cascaded flow inference with feature warping and flow regularization in each pyramid level, which are new in the literature. 
Overall, our network outperforms FlowNet~\cite{Fischer15} and SPyNet~\cite{Ranjan17} and is on par with or outperforms the recent FlowNet2~\cite{Ilg17} on public benchmarks, while having 30 times fewer parameters and being 1.36 times faster than FlowNet2.

\section{Related Work}
\label{sec:related_work}
%
Here, we briefly review some of the major approaches for optical flow estimation.

\vspace{0.1cm}
\noindent \textbf{Variational methods.} Since the pioneering work by Horn and Schunck~\cite{Horn81}, variational methods have dominated optical flow estimation. Brox \etal address illumination changes by combining the brightness and gradient constancy assumptions~\cite{Brox04}. Brox \etal integrate rich descriptors into variational formulation~\cite{Brox11}. In DeepFlow~\cite{Weinzaepfel13}, Weinzaepfel \etal propose to correlate multi-scale patches and incorporate this as the matching term in functional. In PatchMatch Filter~\cite{Lu13}, Lu \etal establish dense correspondence using the superpixel-based PatchMatch~\cite{Barnes09}. Revaud \etal propose a method EpicFlow that uses externally matched flows as initialization and then performs interpolation~\cite{Revaud15}. Zimmer \etal design the complementary regularization that exploits directional information from the constraints imposed in data term~\cite{Zimmer11}. Our network that infers optical flow and performs flow regularization is inspired by the use of data fidelity and regularization in variational methods. 

\vspace{0.1cm}
\noindent \textbf{Machine learning methods.} Black \etal propose to represent complex image motion as a linear combination of the learned basis vectors~\cite{Black97}. Roth \etal formulates the prior probability of flow field as Field-of-Experts model~\cite{Roth05a} that captures higher order spatial statistics~\cite{Roth05b}. Sun \etal study the probabilistic model of brightness inconstancy in a high-order random field framework~\cite{Sun08}. Nir \etal represent image motion using the over-parameterization model~\cite{Nir08}. Rosenbaum \etal model the local statistics of optical flow using Gaussian mixtures~\cite{Rosenbaum13}. Given a set of sparse matches, Wulff \etal propose to regress them to a dense flow field using a set of basis flow fields (PCA-Flow)~\cite{Wulff15}. It can be shown that the parameterized model~\cite{Black97, Nir08, Wulff15} can be efficiently implemented using CNN.

\vspace{0.1cm}
\noindent \textbf{CNN-based methods.} In the work of Fischer \etal termed FlowNet~\cite{Fischer15}, a post-processing step that involves energy minimization is required to reduce smoothing effect across flow boundaries. This process is not end-to-end trainable. In our work, we present an end-to-end approach that performs in-network flow regularization using the proposed f-lconv layer, which plays similar role as the regularization term in variational methods. 
In FlowNet2~\cite{Ilg17}, Ilg \etal introduce a huge network cascade (over 160M parameters) that consists of variants of FlowNet. The cascade improves flow accuracy with an expense of model size and computational complexity. Our model uses a more efficient architecture containing 30 times fewer parameters than FlowNet2 while the performance is on par with it. A compact network termed SPyNet~\cite{Ranjan17} from Ranjan \etal is inspired from spatial pyramid. Nevertheless, the accuracy is far below FlowNet2. A small-sized variant of our network outperforms SPyNet while being 1.33 times smaller in the model size. Zweig \etal present a network to interpolate third-party sparse flows but requiring off-the-shelf edge detector~\cite{Zweig17}. DeepFlow~\cite{Weinzaepfel13} that involves convolution and pooling operations is however not a CNN, since the ``filter weights" are non-trainable image patches. According to the terminology used in FlowNet, DeepFlow uses correlation.

An alternative approach for establishing point correspondence is to match image patches. Zagoruyko \etal first introduce to CNN-feature matching~\cite{Zagoruyko15}. G\"uney \etal find feature representation and formulate optical flow estimation in MRF~\cite{Guney16}. Bailer \etal\cite{Bailer17} use multi-scale features and then perform feature matching as Flow Fields~\cite{Bailer15}. Although pixel-wise matching can establish accurate point correspondence, the computational demand is too high for practical use (it takes several seconds even a GPU is used). As a tradeoff, Fischer \etal~\cite{Fischer15} and Ilg \etal~\cite{Ilg17} perform feature matching only at a reduced spatial resolution. We reduce the computational burden of feature matching by using a short-ranged matching of warped CNN features at sampled positions and a sub-pixel refinement at every pyramid level.

We are inspired by the feature transformation used in Spatial Transformer~\cite{Jaderberg15}. Our network uses the proposed f-warp layer to displace each channel\footnote{We can also use f-warp layer to displace each channel \textit{differently} when multiple flow fields are supplied. The usage, however, is beyond the scope of this work.} of the given vector-valued feature according to the provided flow field. Unlike Spatial Transformer, f-warp layer is not fully constrained and is a relaxed version of it as the flow field is not parameterized. While transformation in FlowNet2 and SPyNet is limited to images, our decider network is a more generic warping network that warps high-level CNN features. 
\begin{figure*}[ht]
\centering
   \includegraphics[width=\linewidth]{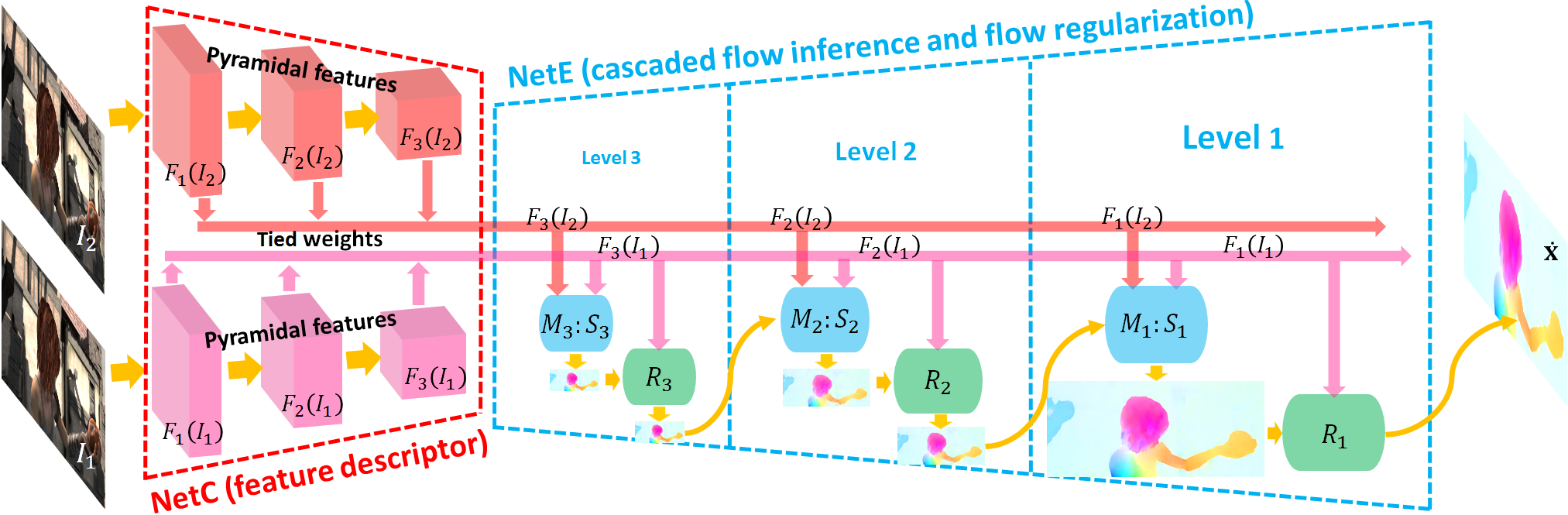}
\caption{The network structure of LiteFlowNet. For the ease of representation, only a 3-level design is shown. Given an image pair ($I_{1}$ and $I_{2}$), NetC generates two pyramids of high-level features ($\mathcal F_{k}(I_{1})$ in pink and $\mathcal F_{k}(I_{2})$ in red, $k \in [1, 3]$). 
NetE yields multi-scale flow fields that each of them is generated by a cascaded flow inference module $M$:$S$ (in blue color, including a descriptor matching unit $M$ and a sub-pixel refinement unit $S$) and a regularization module $R$ (in green color). 
Flow inference and regularization modules correspond to data fidelity and regularization terms in conventional energy minimization methods respectively.}
\label{fig:network structure}
\end{figure*}

\section{LiteFlowNet}
\label{sec:liteflownet}
%
LiteFlowNet is composed of two compact sub-networks that are specialized in pyramidal feature extraction and optical flow estimation as shown in Figure~\ref{fig:network structure}. Since the spatial dimension of feature maps is contracting in feature extraction and that of flow fields is expanding in flow estimation, we call the two sub-networks as NetC and NetE respectively. NetC transforms any given image pair into two pyramids of multi-scale high-dimensional features. NetE consists of cascaded flow inference and regularization modules that estimate coarse-to-fine flow fields. 

\vspace{0.1cm}
\noindent \textbf{Pyramidal Feature Extraction.} As shown in Figure~\ref{fig:network structure}, NetC is a two-stream network in which the filter weights are shared across the two streams. Each of them functions as a feature descriptor that transforms an image $I$ to a pyramid of multi-scale high-dimensional features $\{\mathcal{F}_{k}(I)\}$ from the highest spatial resolution ($k = 1$) to the lowest spatial resolution ($k = L$). The pyramidal features are generated by stride-$s$ convolutions with the reduction of spatial resolution by a factor $s$ up the pyramid. In the following, we omit the subscript $k$ that indicates the level of pyramid for brevity. We use $\mathcal{F}_{i}$ to represent CNN features for $I_{i}$. When we discuss the operations in a pyramid level, the same operations are applicable to other levels. 

\vspace{0.1cm}
\noindent \textbf{Feature Warping.} At each pyramid level, a flow field is inferred from high-level features $\mathcal{F}_{1}$ and ${\mathcal F}_{2}$ of images $I_{1}$ and $I_{2}$. Flow inference becomes more challenging if $I_{1}$ and $I_{2}$ are captured far away from each other. With the motivation of image warping used in conventional methods~\cite{Brox04, Papenberg06} and recent CNNs~\cite{Ilg17, Ranjan17} for addressing large-displacement flow, we propose to reduce feature-space distance between $\mathcal{F}_{1}$ and ${\mathcal F}_{2}$ by feature warping (f-warp). Specifically, ${\mathcal F}_{2}$ is warped towards ${\mathcal F}_{1}$ by f-warp via flow estimate $\dot{\bf x}$ to $\widetilde {\mathcal F}_{2}({\bf x}) \triangleq {\mathcal F}_{2}({\bf x}+\dot{\bf x}) \sim {\mathcal F}_{1}({\bf x})$. This allows our network to infer residual flow between ${\mathcal F}_{1}$ and $\widetilde {\mathcal F}_{2}$ that has smaller flow magnitude (more details in Section~\ref{sec:cascaded flow inference}) but not the complete flow field that is more difficult to infer.
Unlike conventional methods, f-warp is performed on high-level CNN features but not on images. This makes our network more powerful and efficient in addressing the optical flow problem. To allow end-to-end training, ${\mathcal F}$ is interpolated to ${\widetilde {\mathcal F}}$ for any sub-pixel displacement $\dot{\bf x}$ as follows:
\begin{equation}\label{bi interpolation}
\widetilde{\mathcal F}({\bf x}) = \sum_{{\bf x}_{s}^{i} \in N({\bf x}_{s})}{\mathcal F}({\bf x}_{s}^{i})\left(1-\left| x_{s} - x_{s}^{i}\right|\right) \left(1-\left| y_{s} - y_{s}^{i}\right|\right),
\end{equation}
where ${\bf x}_{s} = {\bf x}+\dot{\bf x} = (x_{s}, y_{s})^{\top}$ denotes the source coordinates in the input feature map ${\mathcal F}$ that defines the sample point, ${\bf x} = (x, y)^{\top}$ denotes the target coordinates of the regular grid in the interpolated feature map $\widetilde{\mathcal F}$, and $N({\bf x}_{s})$ denotes the four pixel neighbors of ${\bf x}_{s}$. The above bilinear interpolation allows back-propagation during training as its gradients can be efficiently computed~\cite{Jaderberg15}. 

\subsection{Cascaded Flow Inference}
\label{sec:cascaded flow inference}
%
At each pyramid level of NetE, pixel-by-pixel matching of high-level features yields coarse flow estimate. A subsequent refinement on the coarse flow further improves it to sub-pixel accuracy.

\vspace{0.1cm}
\noindent \textbf{First Flow Inference (descriptor matching).} Point correspondence between $I_{1}$ and $I_{2}$ is established through computing correlation of high-level feature vectors in individual pyramidal features ${\mathcal F}_{1}$ and ${\mathcal F}_{2}$ as follows:
\begin{equation}\label{eq:matching cost}
c({\bf x},{\bf d}) = {\mathcal F}_{1}({\bf x}) \cdot {\mathcal F}_{2}({\bf x}+{\bf d}) / N,
\end{equation}
where $c$ is the matching cost between point ${\bf x}$ in ${\mathcal F}_{1}$ and point ${\bf x}+{\bf d}$ in ${\mathcal F}_{2}$, ${\bf d} \in {\mathbb Z}$ is the displacement vector from ${\bf x}$, and $N$ is the length of the feature vector. A cost volume $C$ is built by aggregating all the matching costs into a 3D grid.

We reduce the computational burden raised by cost-volume processing~\cite{Fischer15, Ilg17} in three ways: 1) We perform short-range matching at every pyramid level instead of long-range matching at a single level. 2) We reduce feature-space distance between ${\mathcal F}_{1}$ and ${\mathcal F}_{2}$ by warping ${\mathcal F}_{2}$ towards ${\mathcal F}_{1}$ using our proposed f-warp through flow estimate\footnote{$\dot{\bf x}$ from previous level needs to be upsampled in spatial resolution (denoted by ``$\uparrow$$s$") and magnitude (multiplied by a scalar $s$) to $s\dot{\bf x}^{\uparrow s}$ for matching the spatial resolution of the pyramidal features at the current level.} $\dot{\bf x}$ from previous level. 3) We perform matching only at the sampled positions in the pyramid levels of high-spatial resolution. The sparse cost volume is interpolated in the spatial dimension to fill the missed matching costs for the unsampled positions. The first two techniques effectively reduce the searching space needed, while the last technique reduces the frequency of matching per pyramid level. 

\begin{figure}[t]
\centering
   \includegraphics[width=\linewidth]{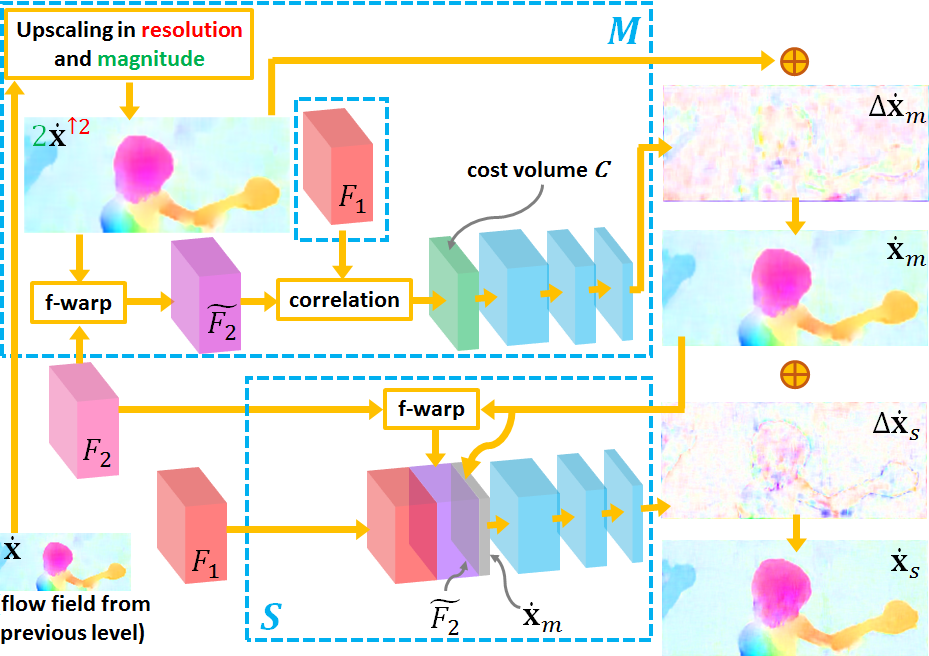}
\caption{A cascaded flow inference module $M$:$S$ in NetE. It consists of a descriptor matching unit $M$ and a sub-pixel refinement unit $S$. In $M$, f-warp transforms high-level feature ${\mathcal F}_{2}$ to $\widetilde{\mathcal F}_{2}$ via upscaled flow field $2 \dot{\bf x}^{ \uparrow 2}$ estimated at previous pyramid level. In $S$, $\mathcal F_{2}$ is warped by $\dot{\bf x}_{m}$ from $M$. In comparison to residual flow $\Delta \dot{\bf x}_{m}$, more flow adjustment exists at flow boundaries in $\Delta \dot{\bf x}_{s}$.}
\label{fig:Ck}
\end{figure}

In the descriptor matching unit $M$, residual flow $\Delta\dot{\bf x}_{m}$ is inferred by filtering the cost volume $C$ as illustrated in Figure~\ref{fig:Ck}. A complete flow field $\dot{\bf x}_{m}$ is computed as follows:
\begin{equation}
\dot{\bf x}_{m} = \underbrace{M\big(C({\mathcal F}_{1}, \widetilde {\mathcal F}_{2}; {\bf d})\big)}_{\Delta \dot{\bf x}_{m}} + s\dot{\bf x}^{\uparrow s}.
\end{equation}

\noindent 
\textbf{Second Flow Inference (sub-pixel refinement).} Since the cost volume in descriptor matching unit is aggregated by measuring pixel-by-pixel correlation, flow estimate $\dot{\bf x}_{m}$ from the previous inference is only up to pixel-level accuracy. We introduce the second flow inference in the wake of descriptor matching as shown in Figure~\ref{fig:Ck}. It aims to refine the pixel-level flow field $\dot{\bf x}_{m}$ to sub-pixel accuracy. This prevents erroneous flows being amplified by upsampling and passing to the next pyramid level. Specifically, ${\mathcal F}_{2}$ is warped to $\widetilde {\mathcal F}_{2}$ via flow estimate $\dot{\bf x}_{m}$. Sub-pixel refinement unit $S$ yields a more accurate flow field $\dot{\bf x}_{s}$ by minimizing feature-space distance between ${\mathcal F}_{1}$ and $\widetilde {\mathcal F}_{2}$ through computing residual flow $\Delta \dot{\bf x}_{s}$ as the following:
\begin{equation}
\dot{\bf x}_{s} = \underbrace{S\big({\mathcal F}_{1}, \widetilde {\mathcal F}_{2}, \dot{\bf x}_{m}\big)}_{\Delta \dot{\bf x}_{s}} + \dot{\bf x}_{m}.
\end{equation}

\subsection{Flow Regularization}
%
Cascaded flow inference resembles the role of data fidelity in conventional minimization methods. Using data term alone, vague flow boundaries and undesired artifacts commonly exist in flow field~\cite{Werlberger09, Zimmer11}. To tackle this problem, we propose to use a feature-driven local convolution (f-lcon) to regularize flow field from the cascaded flow inference. The operation of f-lcon is well-governed by the Laplacian formulation of diffusion of pixel values~\cite{Tschumperle05}.
In contrast to local convolution (lcon) used in conventional CNNs~\cite{Taigman14}, f-lcon is more generalized. Not only is a distinct filter used for each position of feature map, but the filter is adaptively constructed for individual flow patches.

Consider a general case, a vector-valued feature $F$ that has to be regularized has $C$ channels and a spatial dimension $M \times N$. Define ${\bf G} = \{g\}$ as the set of filters used in f-lcon layer. The operation of f-lcon to $F$ can be formulated as follow:
\begin{equation}\label{local convolution1}
  f_{g}(x,y,c) = g(x,y,c) \ast f(x,y,c),
\end{equation}
where ``$\ast$"  denotes convolution,  $f(x,y,c)$ is a $w \times w$ patch centered at position $(x,y)$ of channel $c$ in $F$,  $g(x,y,c)$ is the corresponding $w \times w$ regularization filter, and $f_{g}(x,y,c)$ is a scalar output for ${\bf x} = (x,y)^{\top}$ and $c = 1, 2, ..., C$. To be specific for regularizing flow field $\dot{\bf x}_{s}$ from the cascaded flow inference, we replace $F$ to $\dot{\bf x}_{s}$. Flow regularization module $R$ is defined as follows:
\begin{equation}
     \dot{\bf x}_{r} = R(\dot{\bf x}_{s}; {\bf G}).
\end{equation}

\begin{figure}[t]
\centering
   \includegraphics[width=7.2cm]{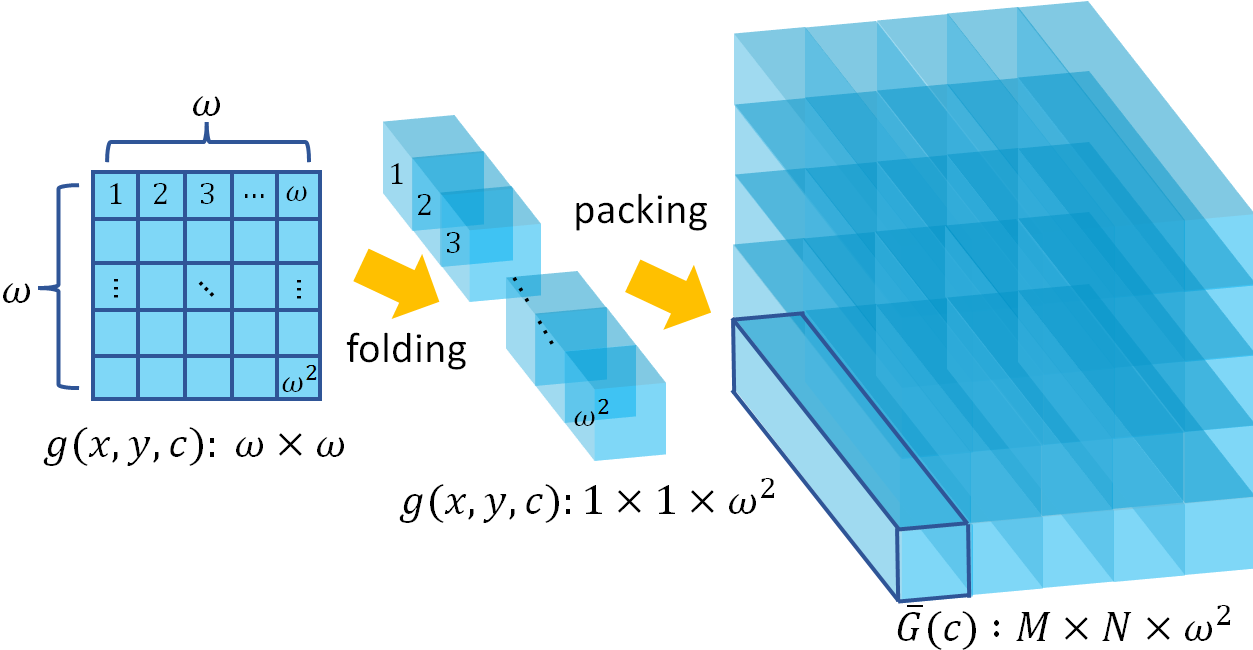}
\caption{Folding and packing of f-lcon filters $\{g\}$. The $(x,y)$-entry of 3D tensor $\bar G(c)$ is a 3D column with size $1 \times 1 \times w^{2}$. It corresponds to the unfolded $w \times w$ f-lcon filter $g(x,y,c)$ to be applied at position $(x,y)$ of channel $c$ in vector-valued feature $F$.}
\label{fig:fold}
\vspace{-1em}
\end{figure}

The f-lcon filters need to be specialized for smoothing flow field. It should behave as an averaging filter if the variation of flow vectors over the patch is smooth. It should also not over-smooth flow field across flow boundary. We define a feature-driven CNN distance metric $\mathcal{D}$ that estimates local flow variation using pyramidal feature $\mathcal{F}_{1}$, flow field $\dot{\bf x}_{s}$ from the cascaded flow inference, and occlusion probability map\footnote{We use the brightness error $||I_{2}({\bf x}+\dot{\bf x})-I_{1}({\bf x})||_{2}$ between the warped second image and the first image as the occlusion probability map.} $O$.
In summary, $\mathcal{D}$ is adaptively constructed by a CNN unit $R_{D}$ as follows:  
\begin{equation}
     \mathcal{D} = R_{D}(\mathcal{F}_{1}, \dot{\bf x}_{s}, O).
\end{equation}
With the introduction of feature-driven distance metric $\mathcal{D}$, each filter $g$ of f-lcon is constructed as follows:
\begin{equation}
     g(x,y,c) = \frac{\text{exp}(-\mathcal{D}(x,y,c)^{2})}{\sum_{(x_{i},y_{i}) \in N(x,y)} \text{exp}(-\mathcal{D}(x_{i},y_{i},c)^{2})},
\end{equation} 
where $N(x,y)$ denotes the neighborhood containing $\omega \times \omega$ pixels centered at position $(x,y)$.

Here, we provide a mechanism to perform f-lcon efficiently. For a $C$-channel input $F$, we use $C$ tensors $\bar G(1), ..., \bar G(C)$ to store f-lcon filter set ${\bf G}$. As illustrated in Figure~\ref{fig:fold}, each f-lcon filter $g(x,y,c)$ is folded into a $1 \times 1 \times w^{2}$ 3D column and then packed into the $(x,y)$-entry of a $M \times N \times w^{2}$ 3D tensor $\bar G(c)$. Same folding and packing operations are also applied to each patch in each channel of $F$. This results $C$ tensors $\bar F(1), ..., \bar F(C)$ for $F$. In this way, Equation~\eqref{local convolution1} can be reformulated to:
\begin{equation}\label{local convolution2}
   F_{g}(c) = \bar G(c) \odot \bar F(c),
\end{equation}
where ``$\odot$" denotes element-wise dot product between the corresponding columns of the tensors. With the abuse of notation, $F_{g}(c)$ means the $c$-th $xy$-slice of the regularized $C$-channel feature $F_{g}$. Equation~\eqref{local convolution2} reduces the dimension of tensors from $M \times N \times w^{2}$ (right-hand side in prior to the dot product) to $M \times N$ (left-hand side).

\section{Experiments}
\label{sec:experiments}
%
\noindent
\textbf{Network Details.} In LiteFlowNet, NetC generates 6-level pyramidal features and NetE predicts flow fields for levels 6 to 2. Flow field in level 2 is upsampled to yield flow field in level~1. 
We set the maximum searching radius in cost-volume to 3 pixels (levels 6 to 4) or 6 pixels (levels 3 to 2). Matching is performed at each position in pyramidal features, except for levels 3 to 2 that it is performed at a regularly sampled grid (a stride of 2). 
All convolution layers use $3\times3$ filters, except each last layer in descriptor matching $M$, sub-pixel refinement $S$, and flow regularization $R$ units uses $5\times5$ (levels 4 to 3) or $7\times7$ (level 2) filters. 
Each convolution layer is followed by a leaky rectified linear unit layer, except f-lcon and the last layer in $M$, $S$ and $R$ CNN units.
More details can be found in the supplementary material.

\vspace{0.1cm}
\noindent
\textbf{Training Details.} We train our network stage-wise by the following steps: 1) NetC and $M_{6}$:$S_{6}$ of NetE is trained for 300k iterations. 2) $R_{6}$ together with the trained network in step 1 is trained for 300k iterations. 3) For levels $k \in [5, 2]$, $M_{k}$:$S_{k}$ followed by $R_{k}$ is added into the trained network each time. The new network cascade is trained for 200k (level 2: 300k) iterations. Filter weights are initialized from previous level. 
Learning rates are initially set to 1e-4, 5e-5, and 4e-5 for levels 6 to 4, 3 and 2 respectively. We reduce it by a factor of 2 starting at 120k, 160k, 200k, and 240k iterations. We use the same loss weight, L2 training loss, Adam optimization, data augmentation (including noise injection), and training schedule~\footnote{We excluded a small amount of training data in Things3D undergoing extremely large flow displacement as advised by the authors~(\url{https://github.com/lmb-freiburg/flownet2/issues}).} (Chairs~\cite{Fischer15}~$\rightarrow$~Things3D~\cite{Mayer16}) as FlowNet2~\cite{Ilg17}. 
We denote \textbf{LiteFlowNet-pre} and \textbf{LiteFlowNet} as the networks trained on Chairs and Chairs~$\rightarrow$~Things3D, respectively. 

\subsection{Results} 
\label{sec:results}
%
We compare several variants of LiteFlowNet to state-of-the-art methods on public benchmarks including FlyingChairs (Chairs)~\cite{Fischer15}, Sintel clean and final~\cite{Butler12}, KITTI12~\cite{Geiger12}, KITTI15~\cite{Menze15}, and Middlebury~\cite{Baker11}. 
\begin{table}[t]
\small
\centering
\caption{AEE on the Chairs testing set. Models are trained on the Chairs training set.} \label{tab:flyingchairs results}
\scalebox{0.82}{
\begin{tabular}{c|c|c|c|c|c}

\hline          
\multicolumn{1}{|c|}{FlowNetS} 							
&\multicolumn{1}{c|}{FlowNetC}						
&\multicolumn{1}{c|}{SPyNet} 
&\multicolumn{1}{c|}{LiteFlowNetX-pre}   
&\multicolumn{1}{c|}{LiteFlowNet-pre} \\

\hline \hline
\multicolumn{1}{|c|}{2.71} 							
&\multicolumn{1}{c|}{2.19}						
&\multicolumn{1}{c|}{2.63} 
&\multicolumn{1}{c|}{2.25}   
&\multicolumn{1}{c|}{\textbf{1.57}} \\
\hline
\end{tabular}}
\end{table}
\begin{table*}[t]
\small
\centering
\caption{AEE of different methods. The values in parentheses are the results of the networks on the data they were trained on, and hence are not directly comparable to the others. Fl-all: Percentage of outliers averaged over all pixels. Inliers are defined as EPE $<$3 pixels or $<$5\%. The best number for each category is highlighted in bold. (Note: $^{1}$The values are reported from~\cite{Ilg17}. $^{2}$We re-trained the model using the code provided by the authors. $^{3,4,5}$The values are computed using the trained models provided by the authors. $^{4}$Large discrepancy exists as the authors mistakenly evaluated the results on the disparity dataset. $^{5}$ Up-to-date dataset is used. $^{6}$Trained on Driving and
Monkaa~\cite{Mayer16})} \label{tab:results}

\scalebox{0.86}{
\begin{tabular}{|c|c||c c|c c|c c|c c c|c c|}
\hline

\multirow{1}{*}{} 
&\multirow{1}{*}{Method}   	                             	
&\multicolumn{2}{c|}{Sintel clean} 							
&\multicolumn{2}{c||}{Sintel final}						
&\multicolumn{2}{c|}{KITTI12} 
&\multicolumn{3}{c||}{KITTI15}   
&\multicolumn{2}{c|}{Middlebury}\\

\multirow{1}{*}{}
&\multirow{1}{*}{}
&\multicolumn{1}{c}{train}&\multicolumn{1}{c|}{test}
&\multicolumn{1}{c}{train}&\multicolumn{1}{c||}{test}
&\multicolumn{1}{c}{train}&\multicolumn{1}{c|}{test}
&\multicolumn{1}{c}{train}&\multicolumn{1}{c}{train (Fl-all)}&\multicolumn{1}{c||}{test (Fl-all)}		
&\multicolumn{1}{c}{train}&\multicolumn{1}{c|}{test}\\	

\hline\hline
\multirow{6}{*}{\rotatebox[origin=c]{90}{Conventional}}
&\multirow{1}{*}{LDOF$^{1}$~\cite{Brox11}}				
&4.64&\multicolumn{1}{c|}{7.56}	           
&5.96&\multicolumn{1}{c||}{9.12}
&10.94&\multicolumn{1}{c|}{12.4}
&18.19&\multicolumn{1}{c}{38.11\%}&\multicolumn{1}{c||}{-}					
&0.44&\multicolumn{1}{c|}{0.56}\\
           
\multirow{1}{*}{}                   
&\multirow{1}{*}{DeepFlow$^{1}$~\cite{Weinzaepfel13}}				
&2.66&\multicolumn{1}{c|}{5.38}	           
&3.57&\multicolumn{1}{c||}{7.21}
&4.48&\multicolumn{1}{c|}{5.8}
&10.63&\multicolumn{1}{c}{26.52\%}&\multicolumn{1}{c||}{29.18\%}		
&0.25&\multicolumn{1}{c|}{0.42}\\
                                                   
\multirow{1}{*}{}                              
&\multirow{1}{*}{Classic+NLP~\cite{Sun14}}				
&4.49&\multicolumn{1}{c|}{6.73}	           
&7.46&\multicolumn{1}{c||}{8.29}
&-&\multicolumn{1}{c|}{7.2}
&-&\multicolumn{1}{c}{-}&\multicolumn{1}{c||}{-}					
&\textbf{0.22}&\multicolumn{1}{c|}{\textbf{0.32}}\\
  
\multirow{1}{*}{}                                                                     
&\multirow{1}{*}{PCA-Layers$^{1}$~\cite{Wulff15}}				
&3.22&\multicolumn{1}{c|}{5.73}	           
&4.52&\multicolumn{1}{c||}{7.89}
&5.99&\multicolumn{1}{c|}{5.2}
&12.74&\multicolumn{1}{c}{27.26\%}&\multicolumn{1}{c||}{-}					
&0.66&\multicolumn{1}{c|}{-}\\                                                                    

\multirow{1}{*}{}
&\multirow{1}{*}{EpicFlow$^{1}$~\cite{Revaud15}}				
&2.27&\multicolumn{1}{c|}{4.12}	           
&3.56&\multicolumn{1}{c||}{6.29}
&\textbf{3.09}&\multicolumn{1}{c|}{3.8}	
&9.27&\multicolumn{1}{c}{27.18\%}&\multicolumn{1}{c||}{\textbf{27.10\%}}				
&0.31&\multicolumn{1}{c|}{0.39}\\
  
\multirow{1}{*}{}                                                                                                                                        
&\multirow{1}{*}{FlowFields$^{1}$~\cite{Bailer15}}				
&\textbf{1.86}&\multicolumn{1}{c|}{\textbf{3.75}}	           
&\textbf{3.06}&\multicolumn{1}{c||}{\textbf{5.81}}
&3.33&\multicolumn{1}{c|}{\textbf{3.5}}	
&\textbf{8.33}&\multicolumn{1}{c}{\textbf{24.43\%}}&\multicolumn{1}{c||}{-}				
&0.27&\multicolumn{1}{c|}{0.33}\\
                                                      
\hline\hline    
\multirow{3}{*}{\rotatebox[origin=c]{90}{Hybrid}}
&\multirow{1}{*}{Deep DiscreteFlow~\cite{Guney16}}				
&-&\multicolumn{1}{c|}{3.86}	           
&-&\multicolumn{1}{c||}{5.73}
&-&\multicolumn{1}{c|}{3.4}	
&-&\multicolumn{1}{c}{-}&\multicolumn{1}{c||}{21.17\%}	
&-&\multicolumn{1}{c|}{-} \\  

\multirow{1}{*}{}
&\multirow{1}{*}{Bailer \etal~\cite{Bailer17}}				
&-&\multicolumn{1}{c|}{\textbf{3.78}}	           
&-&\multicolumn{1}{c||}{5.36}
&-&\multicolumn{1}{c|}{\textbf{3.0}}	
&-&\multicolumn{1}{c}{-}&\multicolumn{1}{c||}{19.44\%}	
&-&\multicolumn{1}{c|}{-} \\  

\multirow{1}{*}{}
&\multirow{1}{*}{DC Flow~\cite{Xu17}}				
&-&\multicolumn{1}{c|}{-}	           
&-&\multicolumn{1}{c||}{\textbf{5.12}}
&-&\multicolumn{1}{c|}{-}	
&-&\multicolumn{1}{c}{-}&\multicolumn{1}{c||}{\textbf{14.86\%}}	
&-&\multicolumn{1}{c|}{-} \\                                                                                                                                                                
                                                                                                                                                                                                                        
\hline\hline    
\multirow{10}{*}{\rotatebox[origin=c]{90}{Heavyweight CNN}}
&\multirow{1}{*}{FlowNetS~\cite{Fischer15}}				
&4.50&\multicolumn{1}{c|}{7.42}	           
&5.45&\multicolumn{1}{c||}{8.43}
&8.26&\multicolumn{1}{c|}{-}
&-&\multicolumn{1}{c}{-}&\multicolumn{1}{c||}{-}			
&1.09&\multicolumn{1}{c|}{-}\\       

\multirow{1}{*}{}
&\multirow{1}{*}{FlowNetS-ft~\cite{Fischer15}}				
&(3.66)&\multicolumn{1}{c|}{6.96}	           
&(4.44)&\multicolumn{1}{c||}{7.76}
&7.52&\multicolumn{1}{c|}{9.1}
&-&\multicolumn{1}{c}{-}&\multicolumn{1}{c||}{-}			
&0.98&\multicolumn{1}{c|}{-} \\ 
                                 
\multirow{1}{*}{}
&\multirow{1}{*}{FlowNetC~\cite{Fischer15}}				
&4.31&\multicolumn{1}{c|}{7.28}	           
&5.87&\multicolumn{1}{c||}{8.81}
&9.35&\multicolumn{1}{c|}{-}	
&-&\multicolumn{1}{c}{-}&\multicolumn{1}{c||}{-}	
&1.15&\multicolumn{1}{c|}{-}\\

\multirow{1}{*}{}
&\multirow{1}{*}{FlowNetC-ft~\cite{Fischer15}}				
&(3.78)&\multicolumn{1}{c|}{6.85}	           
&(5.28)&\multicolumn{1}{c||}{8.51}
&8.79&\multicolumn{1}{c|}{-}	
&-&\multicolumn{1}{c}{-}&\multicolumn{1}{c||}{-}		
&0.93&\multicolumn{1}{c|}{-} \\
  
\multirow{1}{*}{}                                                                  
&\multirow{1}{*}{FlowNet2-S$^{3}$~\cite{Ilg17}}				
&3.79&\multicolumn{1}{c|}{-}	           
&4.99&\multicolumn{1}{c||}{-}
&7.26&\multicolumn{1}{c|}{-}
&14.28&\multicolumn{1}{c}{51.06\%}&\multicolumn{1}{c||}{-}			
&1.04&\multicolumn{1}{c|}{-} \\                                                                                                                                                                                                                                                                               

\multirow{1}{*}{}
&\multirow{1}{*}{\textit{FlowNet2-S re-trained}$^{2}$}				
&\textit{3.96}&\multicolumn{1}{c|}{-}	           
&\textit{5.37}&\multicolumn{1}{c||}{-}
&\textit{7.31}&\multicolumn{1}{c|}{-}
&\textit{14.51}&\multicolumn{1}{c}{\textit{51.38\%}}&\multicolumn{1}{c||}{-}			
&\textit{1.13}&\multicolumn{1}{c|}{-} \\                                                                                                                                                                                                                                                                                                                                                    

\multirow{1}{*}{}
&\multirow{1}{*}{FlowNet2-C$^{3}$~\cite{Ilg17}}				
&3.04&\multicolumn{1}{c|}{-}	           
&4.60&\multicolumn{1}{c||}{-}
&5.79&\multicolumn{1}{c|}{-}
&11.49&\multicolumn{1}{c}{44.09\%}&\multicolumn{1}{c||}{-}			
&0.98&\multicolumn{1}{c|}{-} \\                                                                     

\multirow{1}{*}{}
&\multirow{1}{*}{FlowNet2~\cite{Ilg17}}				
&\textbf{2.02}&\multicolumn{1}{c|}{\textbf{3.96}}	           
&\textbf{3.54}$^{4}$&\multicolumn{1}{c||}{6.02}
&4.01$^{5}$&\multicolumn{1}{c|}{-}
&10.08$^{5}$&\multicolumn{1}{c}{29.99\%$^{5}$}&\multicolumn{1}{c||}{-}			
&\textbf{0.35}&\multicolumn{1}{c|}{\textbf{0.52}} \\ 

\multirow{1}{*}{}                                                                                            
&\multirow{1}{*}{FlowNet2-ft-sintel~\cite{Ilg17}}				
&(1.45)&\multicolumn{1}{c|}{4.16}	           
&(2.19$^{4}$)&\multicolumn{1}{c||}{\textbf{5.74}}
&\textbf{3.54}$^{5}$&\multicolumn{1}{c|}{-}
&\textbf{9.94}$^{5}$&\multicolumn{1}{c}{\textbf{28.02\%}$^{5}$}&\multicolumn{1}{c||}{-}			
&\textbf{0.35}&\multicolumn{1}{c|}{-} \\ 

\multirow{1}{*}{}
&\multirow{1}{*}{FlowNet2-ft-kitti~\cite{Ilg17}}				
&3.43&\multicolumn{1}{c|}{-}	           
&4.83$^{4}$&\multicolumn{1}{c||}{-}
&(1.43$^{5}$)&\multicolumn{1}{c|}{\textbf{1.8}}
&(2.36$^{5}$)&\multicolumn{1}{c}{(8.88\%$^{5}$)}&\multicolumn{1}{c||}{\textbf{11.48\%}}			
&0.56&\multicolumn{1}{c|}{-} \\ 

\hline\hline        
\multirow{7}{*}{\rotatebox[origin=c]{90}{Lightweight CNN}}
\multirow{1}{*}{}
&\multirow{1}{*}{SPyNet~\cite{Ranjan17}}				
&4.12&\multicolumn{1}{c|}{6.69}	           
&5.57&\multicolumn{1}{c||}{8.43}
&9.12&\multicolumn{1}{c|}{-}
&-&\multicolumn{1}{c}{-}&\multicolumn{1}{c||}{-}		
&0.33&\multicolumn{1}{c|}{0.58} \\  

\multirow{1}{*}{}
&\multirow{1}{*}{SPyNet-ft~\cite{Ranjan17}}				
&(3.17)&\multicolumn{1}{c|}{6.64}	           
&(4.32)&\multicolumn{1}{c||}{8.36}
&\textit{3.36}$^{6}$&\multicolumn{1}{c|}{4.1}	
&-&\multicolumn{1}{c}{-}&\multicolumn{1}{c||}{35.07\%}	
&0.33&\multicolumn{1}{c|}{0.58} \\  

\multirow{1}{*}{}
&\multirow{1}{*}{LiteFlowNetX-pre}				
&3.70&\multicolumn{1}{c|}{-}	           
&4.82&\multicolumn{1}{c||}{-}
&6.81&\multicolumn{1}{c|}{-}		
&16.64&\multicolumn{1}{c}{36.64\%}&\multicolumn{1}{c||}{-}		
&0.45&\multicolumn{1}{c|}{-} \\                                                       

\multirow{1}{*}{}
&\multirow{1}{*}{LiteFlowNetX}				
&3.58&\multicolumn{1}{c|}{-}	           
&4.79&\multicolumn{1}{c||}{-}
&6.38&\multicolumn{1}{c|}{-}		
&15.81&\multicolumn{1}{c}{34.90\%}&\multicolumn{1}{c||}{-}		
&0.46&\multicolumn{1}{c|}{-} \\                                                                                                                                                                                                                                                
                                                                
\multirow{1}{*}{}                                                                    
&\multirow{1}{*}{LiteFlowNet-pre}				
&2.78&\multicolumn{1}{c|}{-}	           
&4.17&\multicolumn{1}{c||}{-}
&4.56&\multicolumn{1}{c|}{-}		
&11.58&\multicolumn{1}{c}{32.59\%}&\multicolumn{1}{c||}{-}		
&0.45&\multicolumn{1}{c|}{-} \\                                                                                       

\multirow{1}{*}{}
&\multirow{1}{*}{LiteFlowNet}				
&\textbf{2.48}&\multicolumn{1}{c|}{-}	           
&\textbf{4.04}&\multicolumn{1}{c||}{-}
&\textbf{4.00}&\multicolumn{1}{c|}{-}		
&\textbf{10.39}&\multicolumn{1}{c}{\textbf{28.50\%}}&\multicolumn{1}{c||}{-}		
&0.39&\multicolumn{1}{c|}{-} \\                                                                                      

\multirow{1}{*}{}
&\multirow{1}{*}{LiteFlowNet-ft}				
&(1.35)&\multicolumn{1}{c|}{\textbf{4.54}}           
&(1.78)&\multicolumn{1}{c||}{\textbf{5.38}}
&(1.05)&\multicolumn{1}{c|}{\textbf{1.6}}		
&(1.62)&\multicolumn{1}{c}{(5.58\%)}&\multicolumn{1}{c||}{\textbf{9.38\%}}		
&\textbf{0.30}&\multicolumn{1}{c|}{\textbf{0.40}} \\    
\hline
\end{tabular}}
\end{table*}

\vspace{0.1cm}
\noindent
\textbf{FlyingChairs.} We first compare the intermediate results of different well-performing networks trained on Chairs alone in Table~\ref{tab:flyingchairs results}. Average end-point error (AEE) is reported. LiteFlowNet-pre outperforms the compared networks. No intermediate result is available for FlowNet2~\cite{Ilg17} as each cascade is trained on the Chairs~$\rightarrow$~Things3D schedule individually. Since FlowNetC, FlowNetS (variants of FlowNet~\cite{Fischer15}), and SPyNet~\cite{Ranjan17} have fewer parameters than FlowNet2 and the later two models do not perform feature matching, we also construct a small-size counterpart \textbf{LiteFlowNetX-pre} by removing the matching part and shrinking the model sizes of NetC and NetE by about 4 and 5 times, respectively. Despite that LiteFlowNetX-pre is 43 and 1.33 times smaller than FlowNetC and SPyNet, respectively, it still outperforms these networks and is on par with FlowNetC that uses explicit matching. 

\begin{figure*}[t]
\begin{center}
\captionsetup[subfigure]{labelformat=empty, justification=centering}
\captionsetup[subfloat]{farskip=0pt,captionskip=0pt}
\begin{tabular}{ccccc}
   \includegraphics[width=3.45cm]{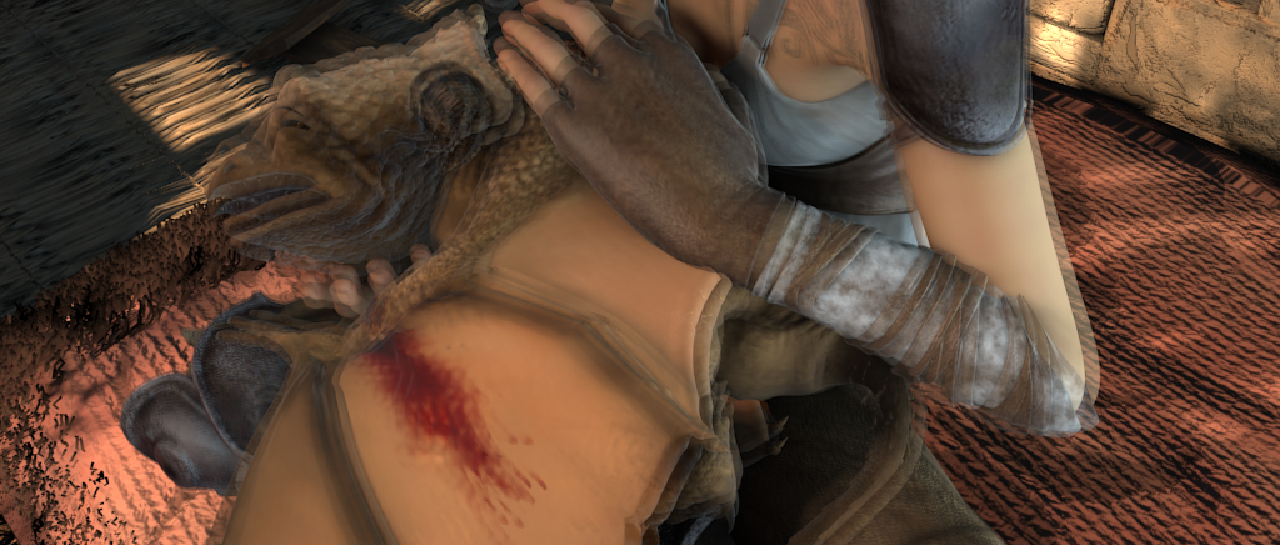}\hfill
   \includegraphics[width=3.45cm]{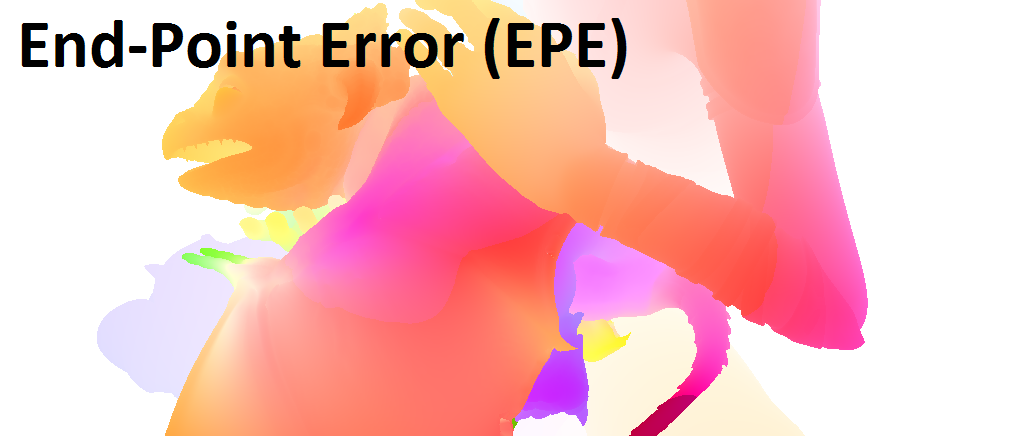}\hfill
   \includegraphics[width=3.45cm]{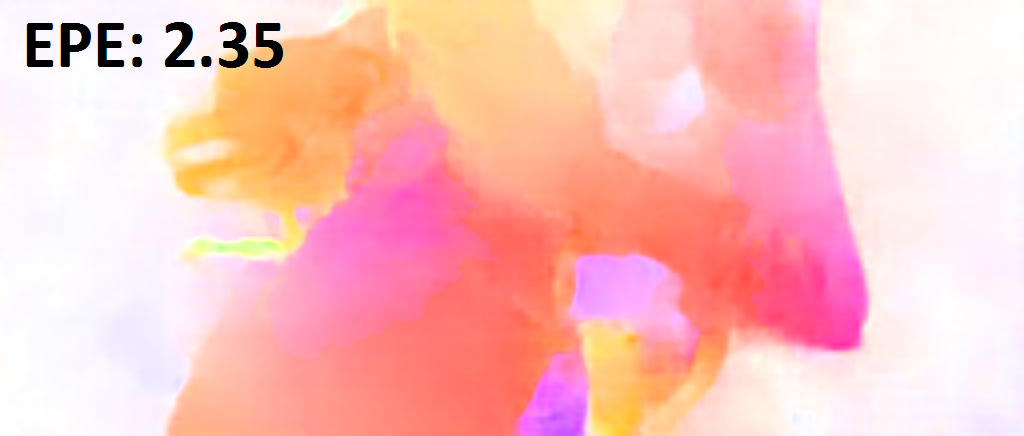}\hfill
   \includegraphics[width=3.45cm]{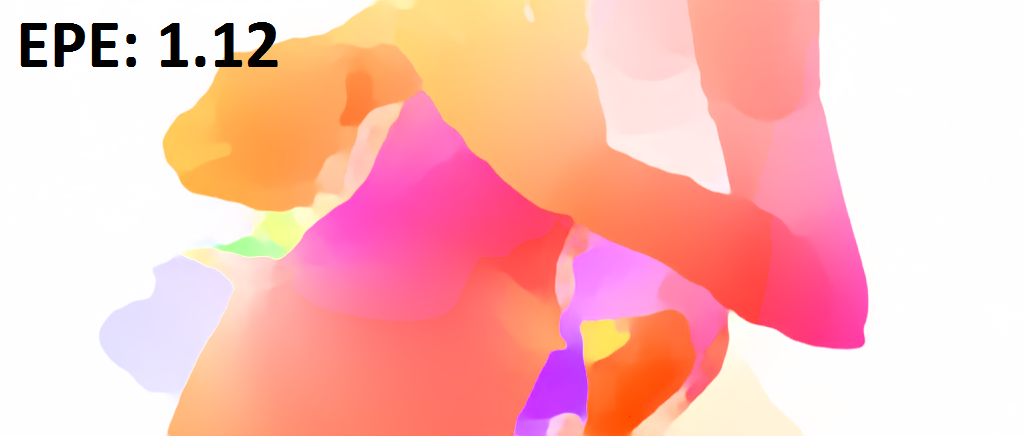}\hfill
   \includegraphics[width=3.45cm]{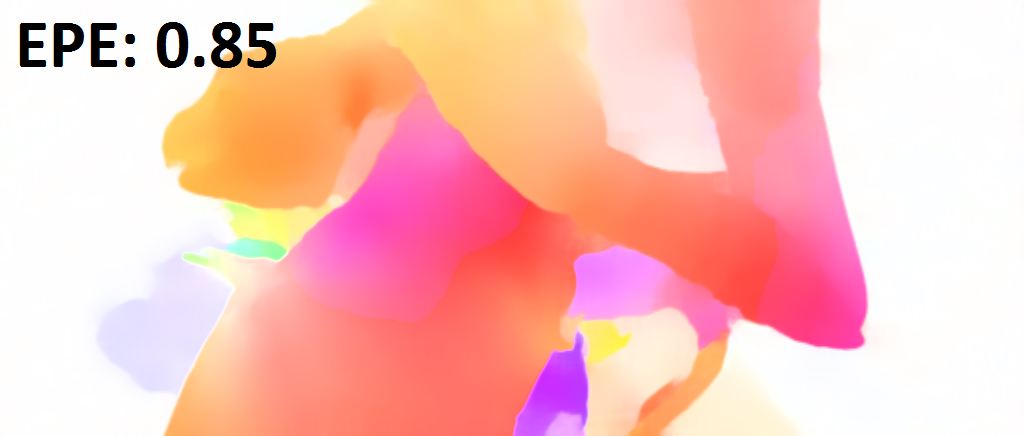} \\
   \subfloat[Image overlay]{\includegraphics[width=3.45cm]{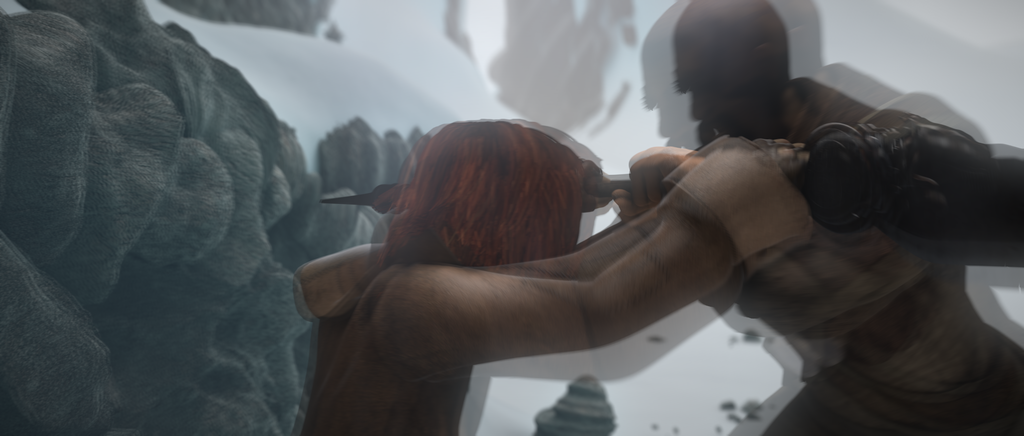}}\hfill
   \subfloat[Ground truth]{\includegraphics[width=3.45cm]{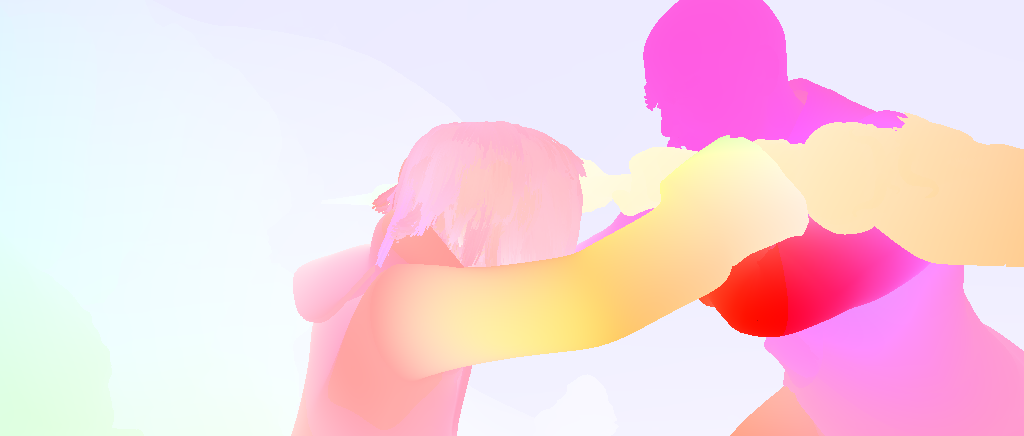}}\hfill
   \subfloat[FlowNetC~\cite{Fischer15}]{\includegraphics[width=3.45cm]{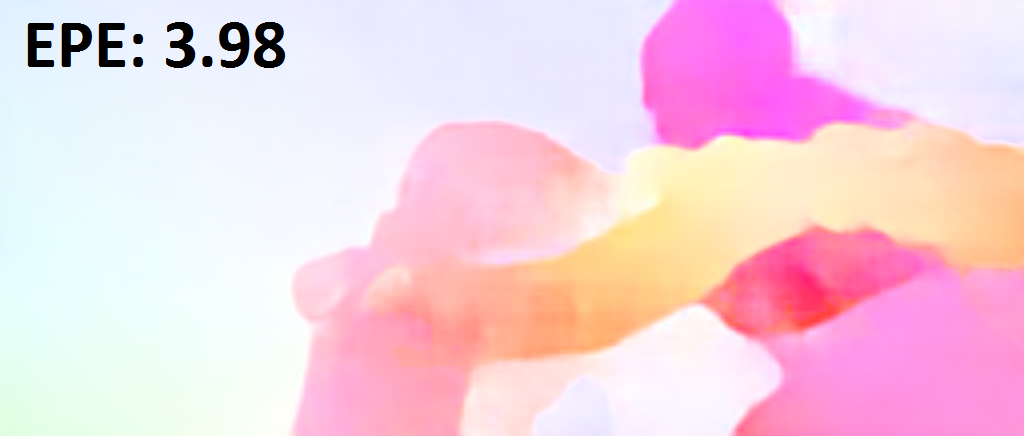}}\hfill
   \subfloat[FlowNet2~\cite{Ilg17}]{\includegraphics[width=3.45cm]{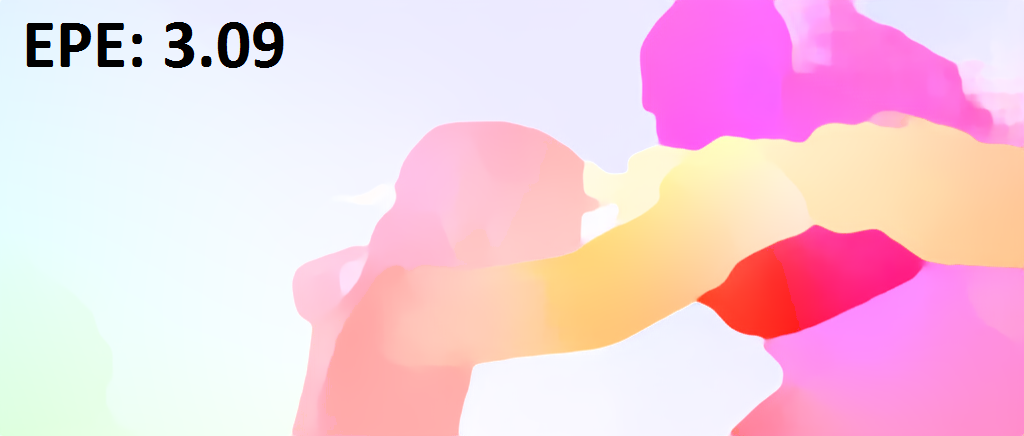}}\hfill
   \subfloat[LiteFlowNet]{\includegraphics[width=3.45cm]{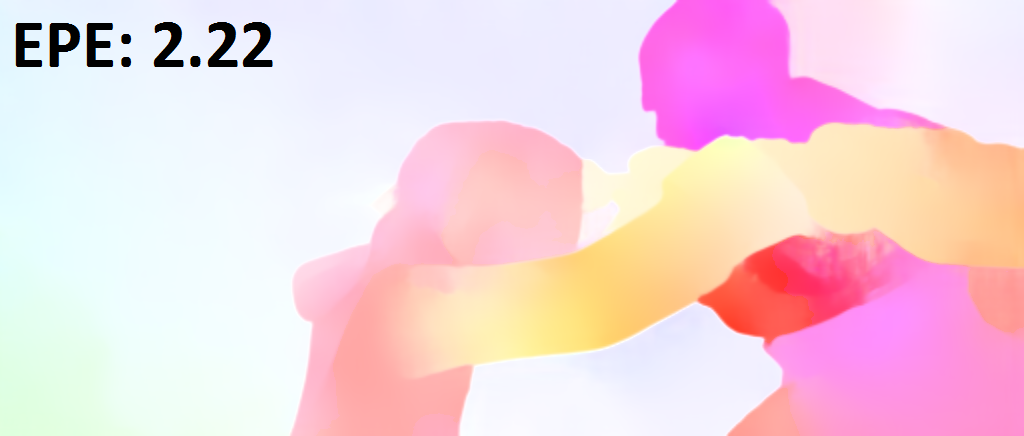}} \\
   \subfloat[First image]{\includegraphics[width=3.45cm]{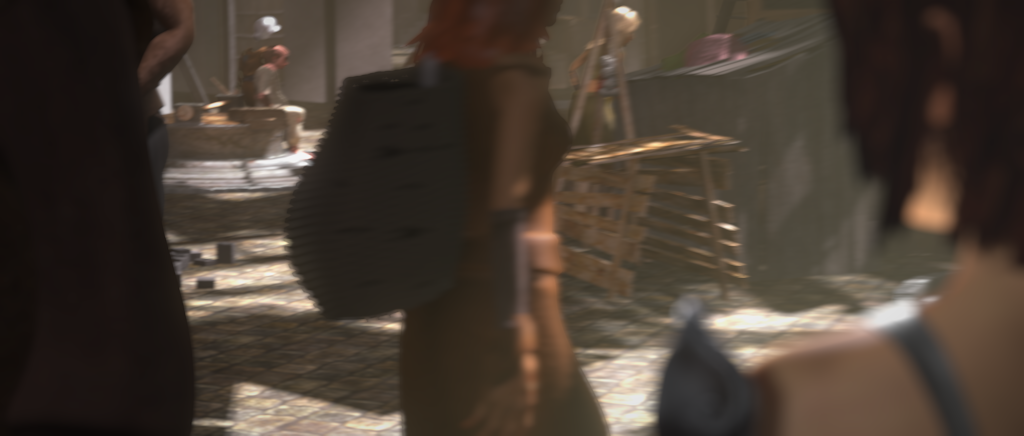}}\hfill
   \subfloat[FlowNetC~\cite{Fischer15}]{\includegraphics[width=3.45cm]{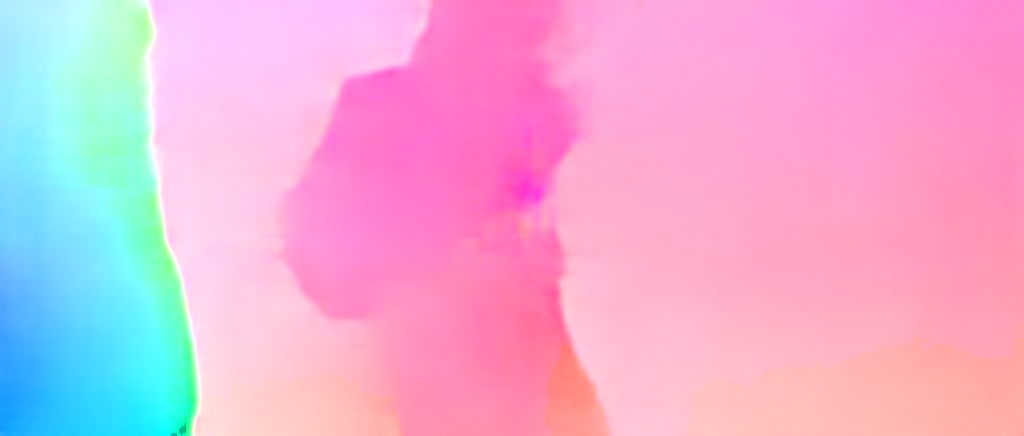}}\hfill
   \subfloat[FlowNet2~\cite{Ilg17}]{\includegraphics[width=3.45cm]{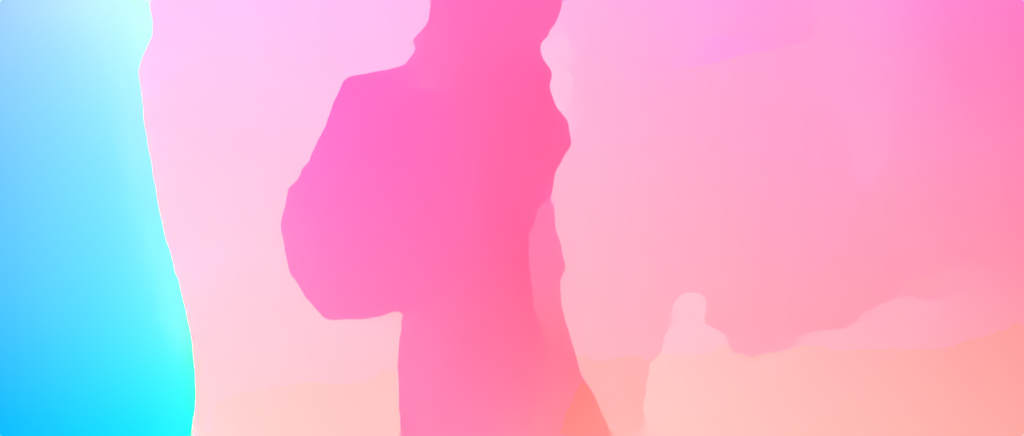}}\hfill
   \subfloat[FlowNet2-ft-sintel~\cite{Ilg17}]{\includegraphics[width=3.45cm]{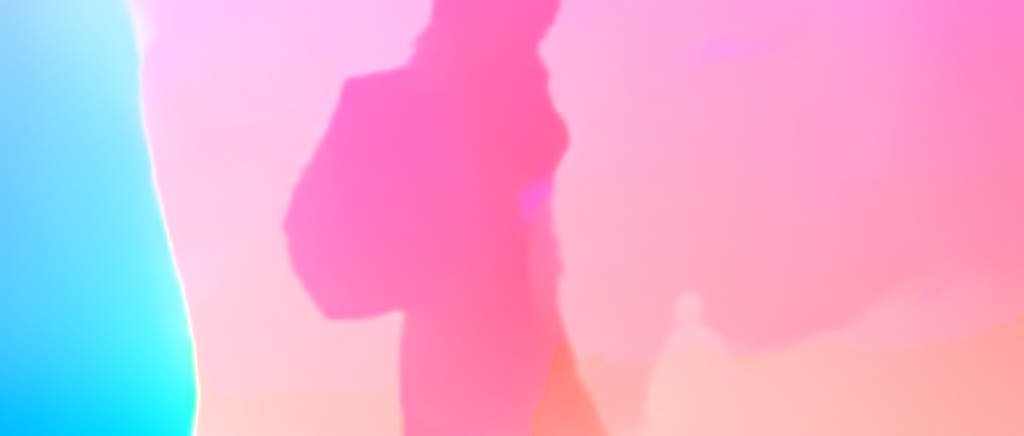}}\hfill
   \subfloat[LiteFlowNet-ft]{\includegraphics[width=3.45cm]{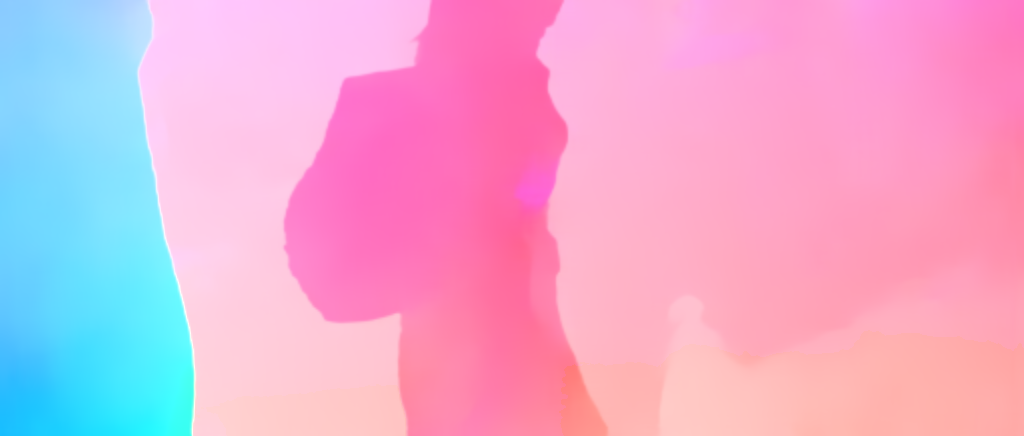}} \\
\end{tabular}
\end{center}
\caption{Examples of flow fields from different methods on Sintel training sets for clean (top row), final (middle row) passes, and the testing set for final pass (last row). Fine details are well preserved and less artifacts can be observed in the flow fields of LiteFlowNet.}
\label{fig:Sintel flows}
\end{figure*}

\vspace{0.1cm}
\noindent
\textbf{MPI Sintel.} In Table~\ref{tab:results}, LiteFlowNetX-pre outperforms FlowNetS (and C)~\cite{Fischer15} and SPyNet~\cite{Ranjan17} that are trained on Chairs on all cases except the Middlebury benchmark. LiteFlowNet, trained on the Chairs~$\rightarrow$~Things3D schedule, performs better than LiteFlowNet-pre as expected. LiteFlowNet also outperforms SPyNet, FlowNet2-S (and -C)~\cite{Ilg17}.
We also fine-tuned LiteFlowNet on a mixture of Sintel clean and final training data (\textbf{LiteFlowNet-ft}) using the generalized Charbonnier loss~\cite{Sun14}. No noise augmentation was performed but we introduced image mirroring to improve the diversity of the training set. LiteFlowNet-ft outperforms FlowNet2-ft-sintel~\cite{Ilg17} and EpicFlow~\cite{Revaud15} for Sintel final testing set. Despite DC Flow~\cite{Xu17} (a hybrid method consists of CNN and post-processing) performs better than LiteFlowNet, its GPU runtime requires several seconds that makes it formidable in many applications. Figure~\ref{fig:Sintel flows} shows some examples of flow fields on Sintel dataset. LiteFlowNet-ft and FlowNet2-ft-sintel perform the best among the compared methods. As LiteFlowNet has flow regularization module, sharper flow boundaries and lesser artifacts can be observed in the generated flow fields. 

\begin{figure*}[t]
\begin{center}
\captionsetup[subfigure]{labelformat=empty, justification=centering}
\captionsetup[subfloat]{farskip=0pt,captionskip=0pt}
\begin{tabular}{ccccc}
   \subfloat[Image overlay]{\includegraphics[width=3.45cm,trim={5cm 0 5cm 0},clip]{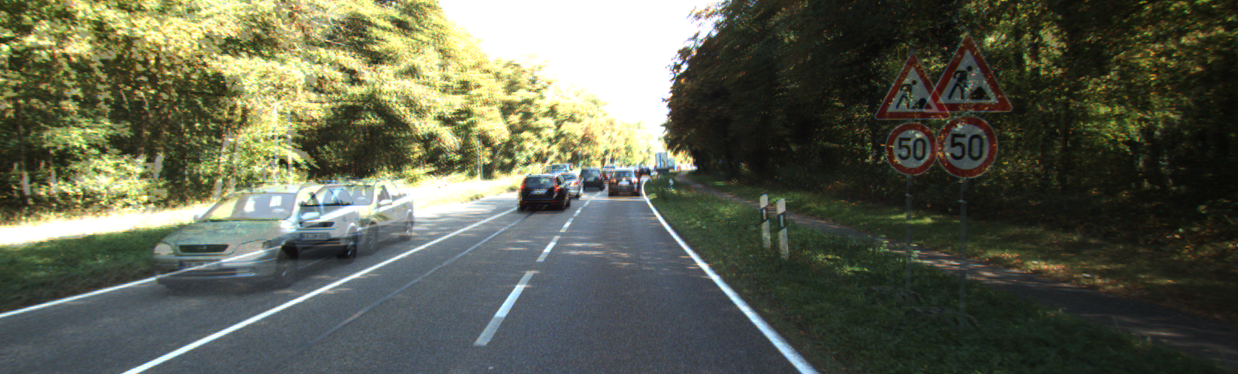}}\hfill
   \subfloat[Ground truth]{\includegraphics[width=3.45cm,trim={5cm 0 5cm 0},clip]{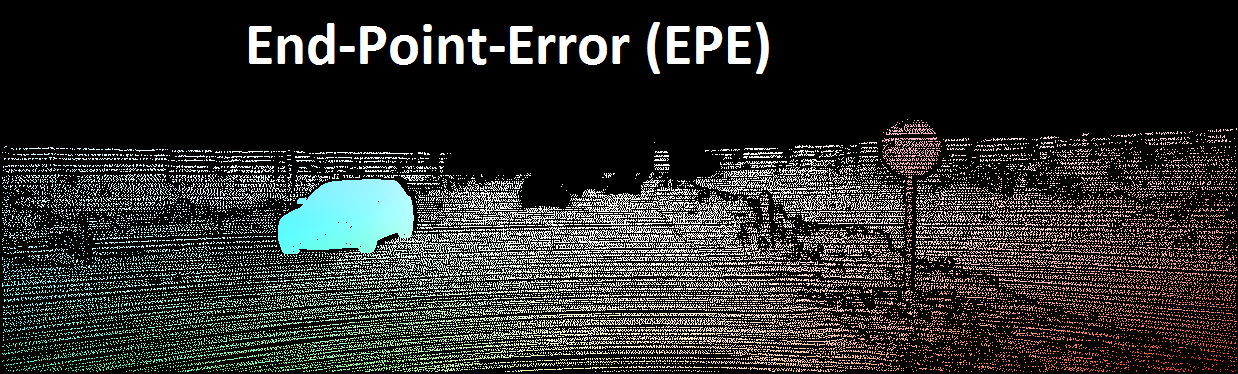}}\hfill
   \subfloat[FlowNetC~\cite{Fischer15}]{\includegraphics[width=3.45cm,trim={5cm 0 5cm 0},clip]{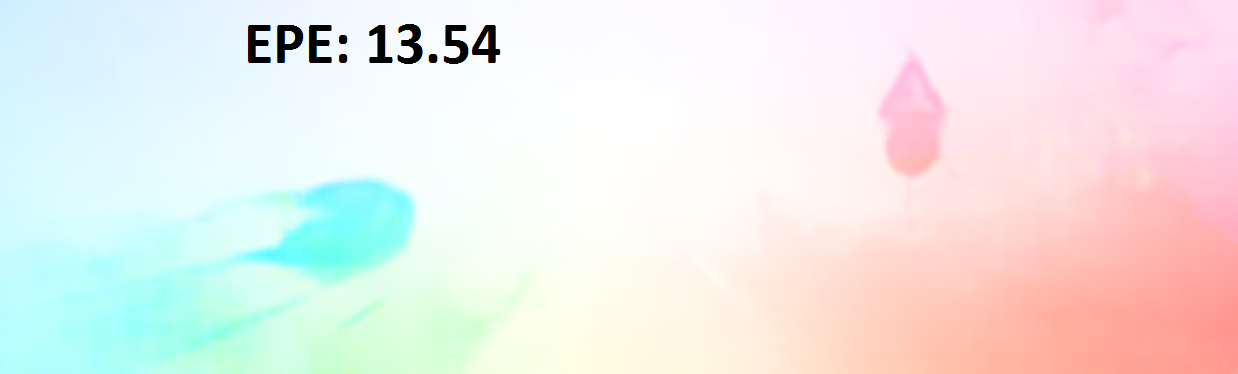}}\hfill
   \subfloat[FlowNet2~\cite{Ilg17}]{\includegraphics[width=3.45cm,trim={5cm 0 5cm 0},clip]{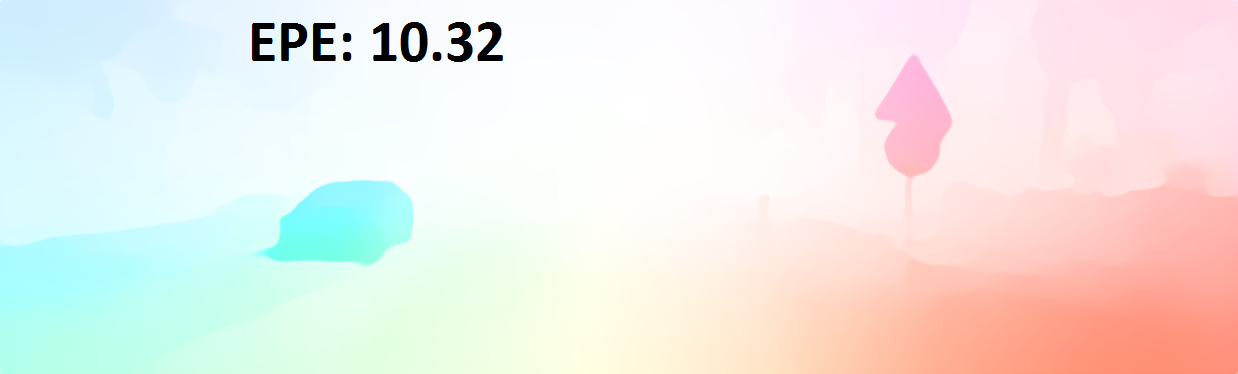}}\hfill
   \subfloat[LiteFlowNet]{\includegraphics[width=3.45cm,trim={5cm 0 5cm 0},clip]{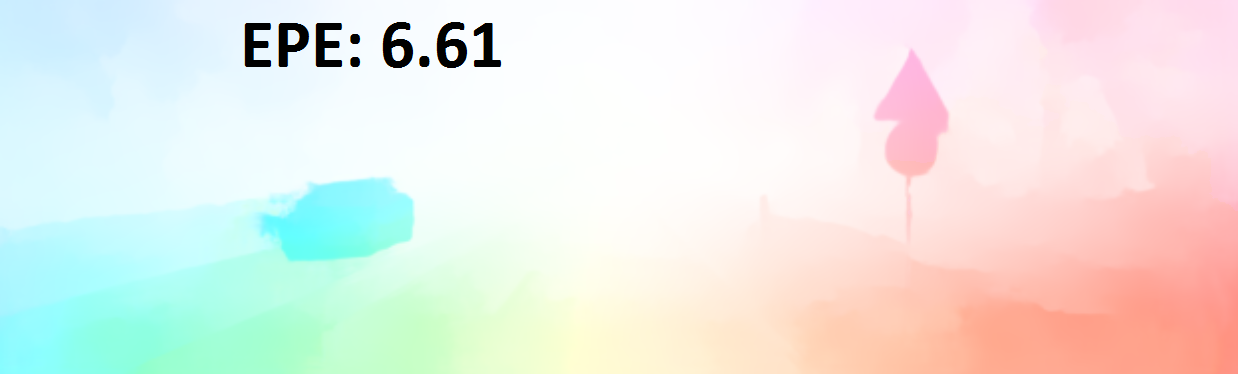}} \\
   \subfloat[First Image]{\includegraphics[width=3.45cm,trim={5cm 0 5cm 0},clip]{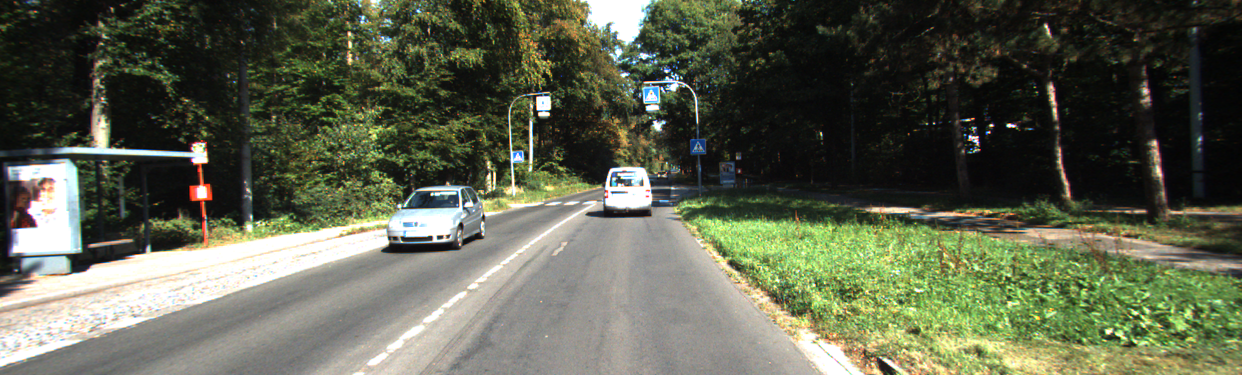}}\hfill
   \subfloat[FlowNetC~\cite{Fischer15}]{\includegraphics[width=3.45cm,trim={5cm 0 5cm 0},clip]{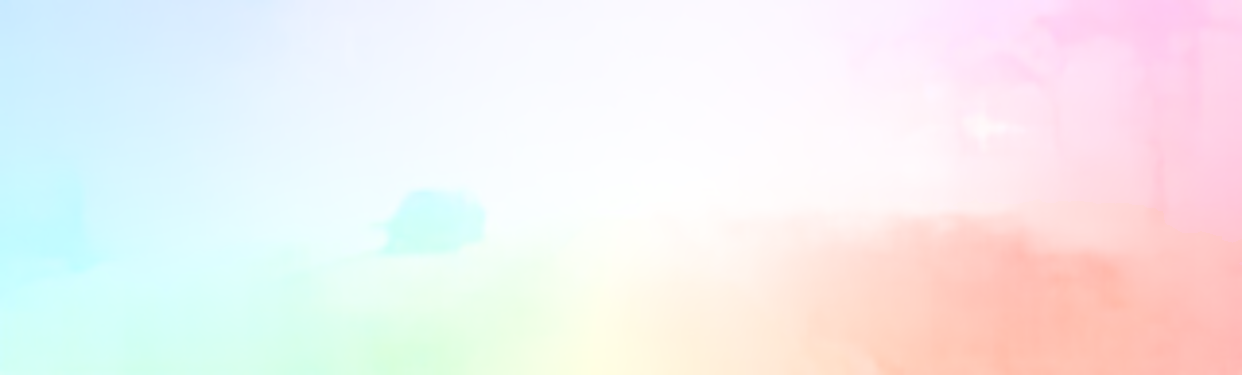}}\hfill
   \subfloat[FlowNet2~\cite{Ilg17}]{\includegraphics[width=3.45cm,trim={5cm 0 5cm 0},clip]{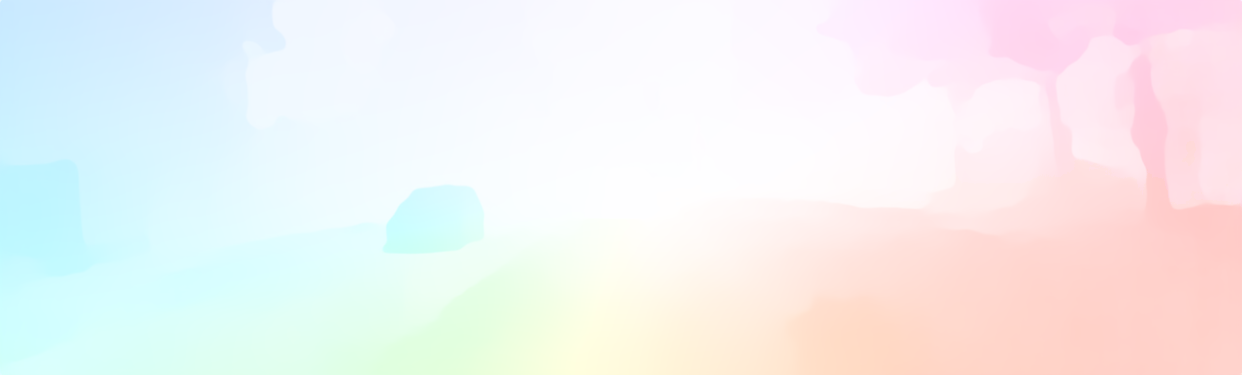}}\hfill
   \subfloat[FlowNet2-ft-kitti~\cite{Ilg17}]{\includegraphics[width=3.45cm,trim={5cm 0 5cm 0},clip]{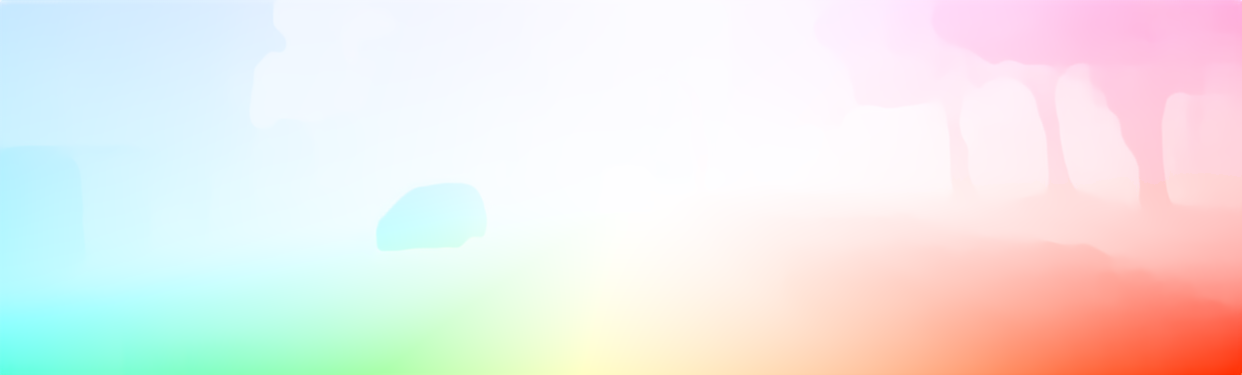}}\hfill
   \subfloat[LiteFlowNet-ft]{\includegraphics[width=3.45cm,trim={5cm 0 5cm 0},clip]{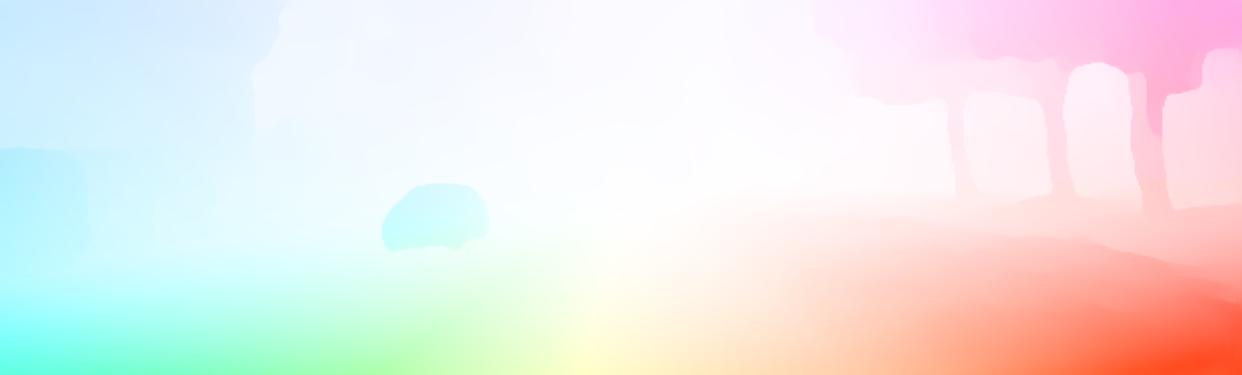}} \\
\end{tabular}
\end{center}
\caption{Examples of flow fields from different methods on the training set (top) and the testing set (bottom) of KITTI15.}
\label{fig:KITTI flows}
\end{figure*}

\vspace{0.1cm}
\noindent
\textbf{KITTI.} LiteFlowNet consistently performs better than LiteFlowNet-pre especially on KITTI15  as shown in Table~\ref{tab:results}. It also outperforms SPyNet~\cite{Ranjan17} and FlowNet2-S (and C)~\cite{Ilg17}. 
We also fine-tuned LiteFlowNet on a mixture of KITTI12 and KITTI15 training data (\textbf{LiteFlowNet-ft}) using the same augmentation as the case of Sintel except that we reduced the amount of augmentation for spatial motion to fit the driving scene. After fine-tuning, LiteFlowNet generalizes well to real-world data. LiteFlowNet-ft outperforms FlowNet2-ft-kitti~\cite{Ilg17}. Figure~\ref{fig:KITTI flows} shows some examples of flow fields on KITTI. As in the case for Sintel, LiteFlowNet-ft and FlowNet2-ft-kitti performs the best among the compared methods. Even though LiteFlowNet and its variants perform pyramidal descriptor matching in a limited searching range, it yields reliable large-displacement flow fields for real-world data due to the feature warping (f-warp) layer introduced. More analysis will be presented in Section~\ref{sec:ablation study}.

\vspace{0.1cm}
\noindent
\textbf{Middlebury.} LiteFlowNet has comparable performance with conventional methods. It outperforms FlowNetS (and C)~\cite{Fischer15}, FlowNet2-S (and C)~\cite{Ilg17},  SPyNet~\cite{Ranjan17}, and FlowNet2~\cite{Ilg17}. On the benchmark, LiteFlowNet-ft refers to the one fine-tuned on Sintel.
\begin{figure*}[ht]
\centering
   \includegraphics[width=4cm]{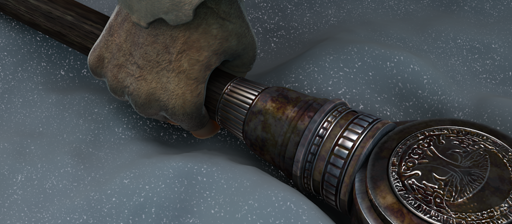}
   \includegraphics[width=4cm]{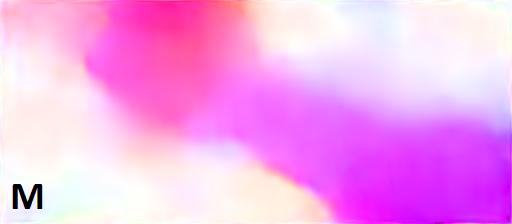} 
   \includegraphics[width=4cm]{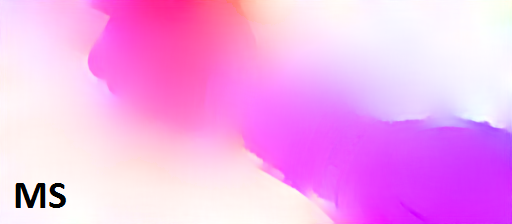} 
   \includegraphics[width=4cm]{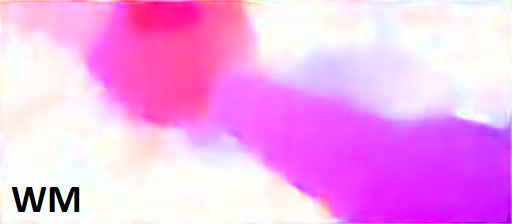}\\
   \includegraphics[width=4cm]{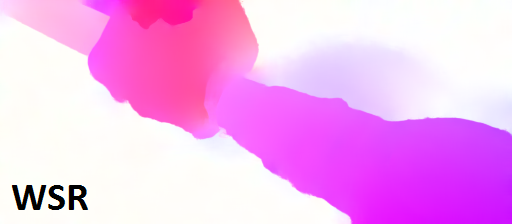} 
   \includegraphics[width=4cm]{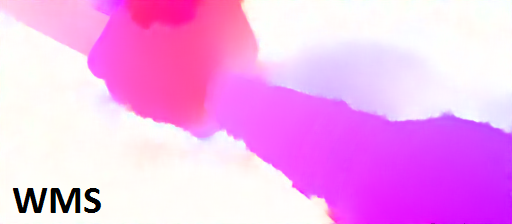}
   \includegraphics[width=4cm]{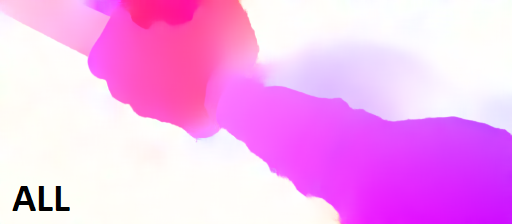}
   \includegraphics[width=4cm]{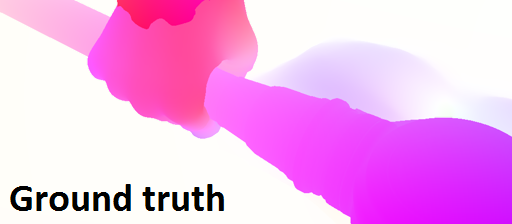}
\caption{Examples of flow fields from different variants of LiteFlowNet-pre trained on Chairs with some of the components disabled. LiteFlowNet-pre is denoted as ``All''. W $=$ Feature \textbf{W}arping, M $=$ Descriptor \textbf{M}atching, S $=$ \textbf{S}ub-Pixel Refinement, R $=$ \textbf{R}egularization.}
\label{fig:ablation flows}
\end{figure*}

\subsection{Runtime and Parameters}
\label{sec:runtime}
\begin{table}[t]
\small
\centering
\caption{Number of training parameters and runtime. The model for which the runtime is in parentheses is measured using Torch, and hence are not directly comparable to the others using Caffe. Abbreviation LFlowNet refers to LiteFlowNet.} \label{tab:model size and runtime}
\scalebox{0.77}{
\begin{tabular}{cccccc}
\hline

\multicolumn{1}{|c||}{}  				&\multicolumn{2}{c||}{Shallow}        
											&\multicolumn{1}{c||}{Deep} 
											&\multicolumn{2}{c|}{Very Deep}  \\ 
\hline
\multicolumn{1}{|c||}{Model} 		     &\multicolumn{1}{c|}{FlowNetC}
											&\multicolumn{1}{c||}{SPyNet}
											&\multicolumn{1}{c||}{LFlowNetX}
											&\multicolumn{1}{c|}{LFlowNet}
											&\multicolumn{1}{c|}{FlowNet2} \\
\hline\hline
\multicolumn{1}{|c||}{\# layers}		&\multicolumn{1}{c|}{26}
											&\multicolumn{1}{c||}{35}
											&\multicolumn{1}{c||}{74}
											&\multicolumn{1}{c|}{99} 
											&\multicolumn{1}{c|}{115} \\
																		
\multicolumn{1}{|c||}{\# param. (M)}	&\multicolumn{1}{c|}{39.16}
											&\multicolumn{1}{c||}{\textbf{1.20}}
											&\multicolumn{1}{c||}{\textbf{0.90}}
											&\multicolumn{1}{c|}{\textbf{5.37}}
											&\multicolumn{1}{c|}{162.49} \\
											 
\multicolumn{1}{|c||}{Runtime (ms)} 	&\multicolumn{1}{c|}{\textbf{32.28}}
											&\multicolumn{1}{c||}{(129.83)}
											&\multicolumn{1}{c||}{\textbf{35.83}}
											&\multicolumn{1}{c|}{\textbf{90.25}}
											&\multicolumn{1}{c|}{122.39} \\
											
\hline
\end{tabular}}
\end{table}
We measure runtime of a CNN using a machine equipped with an Intel Xeon E5 2.2GHz and an NVIDIA GTX 1080. Timings are averaged over 100 runs for Sintel image pairs of size $1024\times436$. As summarized in Table~\ref{tab:model size and runtime}, LiteFlowNet has about \textbf{30 times} fewer parameters than FlowNet2~\cite{Ilg17} and is \textbf{1.36 times} faster in runtime. LiteFlowNetX, a variant of LiteFlowNet having a smaller model size and without descriptor matching, has about \textbf{43 times} fewer parameters than FlowNetC~\cite{Fischer15} and a comparable runtime. LiteFlowNetX also has \textbf{1.33 times} fewer parameters than SPyNet~\cite{Ranjan17}. LiteFlowNet and its variants are currently the most compact CNNs for flow estimation.
%
\subsection{Ablation Study}
\label{sec:ablation study}
\begin{table}[t]
\small
\centering
\caption{AEE of different variants of LiteFlowNet-pre trained on Chairs dataset with some of the components disabled.} \label{tab:ablation study}
\scalebox{0.8}{
\begin{tabular}{|c|c|c|c|c|c|c|}
\hline
                           
\multicolumn{1}{|c||}{Variants}
&\multicolumn{1}{c|}{M}
&\multicolumn{1}{c|}{MS}
&\multicolumn{1}{c|}{WM}
&\multicolumn{1}{c|}{WSR}
&\multicolumn{1}{c||}{WMS}
&\multicolumn{1}{c|}{ALL}  \\
\hline \hline

\multicolumn{1}{|c||}{Feature \textbf{W}arping}
&\multicolumn{1}{c|}{\xmark}
&\multicolumn{1}{c|}{\xmark}
&\multicolumn{1}{c|}{\cmark}
&\multicolumn{1}{c|}{\cmark}
&\multicolumn{1}{c||}{\cmark}
&\multicolumn{1}{c|}{\cmark} \\

\multicolumn{1}{|c||}{Descriptor \textbf{M}atching}
&\multicolumn{1}{c|}{\cmark}
&\multicolumn{1}{c|}{\cmark}
&\multicolumn{1}{c|}{\cmark}
&\multicolumn{1}{c|}{\xmark}
&\multicolumn{1}{c||}{\cmark}
&\multicolumn{1}{c|}{\cmark} \\

\multicolumn{1}{|c||}{\textbf{S}ub-pix. Refinement}
&\multicolumn{1}{c|}{\xmark}
&\multicolumn{1}{c|}{\cmark}
&\multicolumn{1}{c|}{\xmark}
&\multicolumn{1}{c|}{\cmark}
&\multicolumn{1}{c||}{\cmark}
&\multicolumn{1}{c|}{\cmark} \\

\multicolumn{1}{|c||}{\textbf{R}egularization}
&\multicolumn{1}{c|}{\xmark}
&\multicolumn{1}{c|}{\xmark}
&\multicolumn{1}{c|}{\xmark}
&\multicolumn{1}{c|}{\cmark}
&\multicolumn{1}{c||}{\xmark}
&\multicolumn{1}{c|}{\cmark} \\
\hline \hline 

\multicolumn{1}{|c||}{FlyingChairs (train)}
&\multicolumn{1}{c|}{3.75}
&\multicolumn{1}{c|}{2.70}
&\multicolumn{1}{c|}{2.98}
&\multicolumn{1}{c|}{1.63}
&\multicolumn{1}{c||}{1.82}
&\multicolumn{1}{c|}{\textbf{1.57}} \\

\multicolumn{1}{|c||}{Sintel clean (train)}
&\multicolumn{1}{c|}{4.70}
&\multicolumn{1}{c|}{4.17}
&\multicolumn{1}{c|}{3.54}
&\multicolumn{1}{c|}{3.19}
&\multicolumn{1}{c||}{2.90}
&\multicolumn{1}{c|}{\textbf{2.78}} \\

\multicolumn{1}{|c||}{Sintel final (train)}
&\multicolumn{1}{c|}{5.69}
&\multicolumn{1}{c|}{5.30}
&\multicolumn{1}{c|}{4.81}
&\multicolumn{1}{c|}{4.63}
&\multicolumn{1}{c||}{4.45}
&\multicolumn{1}{c|}{\textbf{4.17}} \\

\multicolumn{1}{|c||}{KITTI12 (train)}
&\multicolumn{1}{c|}{9.22}
&\multicolumn{1}{c|}{8.01}
&\multicolumn{1}{c|}{6.17}
&\multicolumn{1}{c|}{5.03}
&\multicolumn{1}{c||}{4.83}
&\multicolumn{1}{c|}{\textbf{4.56}} \\

\multicolumn{1}{|c||}{KITTI15 (train)}
&\multicolumn{1}{c|}{18.24}
&\multicolumn{1}{c|}{16.19}
&\multicolumn{1}{c|}{14.52}
&\multicolumn{1}{c|}{13.20}
&\multicolumn{1}{c||}{12.32}
&\multicolumn{1}{c|}{\textbf{11.58}} \\
\hline 
\end{tabular}}
\end{table}
We investigate the role of each component in LiteFlowNet-pre trained on Chairs by evaluating the performance of different variants with some of the components disabled. The AEE results are summarized in Table~\ref{tab:ablation study} and examples of flow fields are illustrated in Figure~\ref{fig:ablation flows}. 

\vspace{0.1cm}
\noindent
\textbf{Feature Warping.} We consider two variants LiteFlowNet-pre (WM and WMS) and compare them to the counterparts with warping disabled (M and MS). Flow fields from M and MS are more vague. Large degradation in AEE is noticed especially for KITTI12 ($33\%$) and KITTI15 ($25\%$). With feature warping, pyramidal features that input to flow inference are closer to each other. This facilitates flow estimation in subsequent pyramid level by computing residual flow.

\vspace{0.1cm}
\noindent
\textbf{Descriptor Matching.} We compare the variant WSR without descriptor matching for which the flow inference part is made as deep as that in the unamended LiteFlowNet-pre (ALL). No noticeable difference between the flow fields from WSR and ALL. Since the maximum displacement of the example flow field is not very large (only 14.7 pixels), accurate flow field can still be yielded from WSR. For evaluation covering a wide range of flow displacement (especially large-displacement benchmark, KITTI), degradation in AEE is noticed for WSR. This suggests that descriptor matching is useful in addressing large-displacement flow.   

\vspace{0.1cm}
\noindent
\textbf{Sub-Pixel Refinement.} The flow field generated from WMS is more crisp and contains more fine details than that generated from WM with sub-pixel refinement disabled. Less small-magnitude flow artifacts (represented by light color on the background) are also observed. Besides, WMS achieves smaller AEE. Since descriptor matching establishes pixel-by-pixel correspondence, sub-pixel refinement is necessary to yield detail-preserving flow field.

\vspace{0.1cm}
\noindent
\textbf{Regularization.} In comparison WMS with regularization disabled to ALL, undesired artifacts exist in homogeneous regions (represented by very dim color on the background) of the flow field generated from WMS. Flow bleeding and vague flow boundaries are observed. Degradation in AEE is also noticed. This suggests that the proposed feature-driven local convolution (f-lcon) plays the vital role to smooth flow field and maintain crisp flow boundaries as regularization term in conventional variational methods.

\section{Conclusion}
\label{sec:conclusions}
We have presented a compact network for accurate flow estimation. LiteFlowNet outperforms FlowNet~\cite{Fischer15} and is on par with or outperforms the state-of-the-art FlowNet2~\cite{Ilg17} on public benchmarks while being faster in runtime and 30 times smaller in model size. Pyramidal feature extraction and feature warping (f-warp) help us to break the de facto rule of accurate flow network requiring large model size. To address large-displacement and detail-preserving flows, LiteFlowNet exploits short-range matching to generate pixel-level flow field and further improves the estimate to sub-pixel accuracy in the cascaded flow inference. To result crisp flow boundaries, LiteFlowNet regularizes flow field through feature-driven local convolution (f-lcon). With its lightweight, accurate, and fast flow computation, we expect that LiteFlowNet can be deployed to many applications such as motion segmentation, action recognition, SLAM, 3D reconstruction and more.

\vspace{0.1cm}
\noindent\textbf{Acknowledgement.}
This work is supported by SenseTime Group Limited and the General Research Fund sponsored by the Research Grants Council of the Hong Kong SAR (CUHK 14241716, 14224316, 14209217). 

\section{Appendix}
LiteFlowNet consists of two compact sub-networks, namely NetC and NetE. NetC is a two-steam network in which the two network streams share the same set of filters. The input to NetC is an image pair ($I_{1}$, $I_{2}$). The network architectures of the 6-level NetC and NetE at pyramid level 5 are provided in Table \ref{tab:NetC} and Tables \ref{tab:NetE-M-L5} to \ref{tab:NetE-R-L5}, respectively. We use suffixes ``M'', ``S'' and ``R'' to highlight the layers that are used in descriptor matching, sub-pixel refinement, and flow regularization units in NetE, respectively. We declare a layer as ``flow'' to highlight when the output is a flow field. Our code and trained models are available at \url{https://github.com/twhui/LiteFlowNet}. A video clip (\url{https://www.youtube.com/watch?v=pfQ0zFwv-hM}) and a supplementary material are available on our project page (\url{http://mmlab.ie.cuhk.edu.hk/projects/LiteFlowNet/}) to showcase the performance of LiteFlowNet and the effectiveness of the proposed components in our network. 

\begin{table*}[t]
\centering
\caption{The network details of NetC in LiteFlowNet. ``\# Ch. In / Out'' means the number of channels of the input or the output features. ``conv'' denotes convolution.} \label{tab:NetC}
\scalebox{1}{
\begin{tabular}{cccccc}
\hline

\multicolumn{1}{|c||}{Layer name} 			
&\multicolumn{1}{c|}{Kernel}
&\multicolumn{1}{c|}{Stride}									
&\multicolumn{1}{c|}{\# Ch. In / Out}
&\multicolumn{1}{c|}{Input} \\
\hline\hline

\multicolumn{1}{|c||}{conv1}			
&\multicolumn{1}{c|}{7$\times$7}
&\multicolumn{1}{c|}{1}
&\multicolumn{1}{c|}{3 / 32} 
&\multicolumn{1}{c|}{$I_{1}$ or $I_{2}$} \\
																		
\multicolumn{1}{|c||}{conv2\_1}			
&\multicolumn{1}{c|}{3$\times$3}
&\multicolumn{1}{c|}{2}
&\multicolumn{1}{c|}{32 / 32}
&\multicolumn{1}{c|}{conv1} \\
											 
\multicolumn{1}{|c||}{conv2\_2}			
&\multicolumn{1}{c|}{3$\times$3}
&\multicolumn{1}{c|}{1}
&\multicolumn{1}{c|}{32 / 32}
&\multicolumn{1}{c|}{conv2\_1} \\
		
\multicolumn{1}{|c||}{conv2\_3}			
&\multicolumn{1}{c|}{3$\times$3}
&\multicolumn{1}{c|}{1}
&\multicolumn{1}{c|}{32 / 32}
&\multicolumn{1}{c|}{conv2\_2} \\

\multicolumn{1}{|c||}{conv3\_1}			
&\multicolumn{1}{c|}{3$\times$3}
&\multicolumn{1}{c|}{2}
&\multicolumn{1}{c|}{32 / 64}
&\multicolumn{1}{c|}{conv2\_3} \\																							
\multicolumn{1}{|c||}{conv3\_2}			
&\multicolumn{1}{c|}{3$\times$3}
&\multicolumn{1}{c|}{1}
&\multicolumn{1}{c|}{64 / 64}
&\multicolumn{1}{c|}{conv3\_1} \\

\multicolumn{1}{|c||}{conv4\_1}			
&\multicolumn{1}{c|}{3$\times$3}
&\multicolumn{1}{c|}{2}
&\multicolumn{1}{c|}{64 / 96}
&\multicolumn{1}{c|}{conv3\_2} \\																							
\multicolumn{1}{|c||}{conv4\_2}			
&\multicolumn{1}{c|}{3$\times$3}
&\multicolumn{1}{c|}{1}
&\multicolumn{1}{c|}{96 / 96}
&\multicolumn{1}{c|}{conv4\_1} \\

\multicolumn{1}{|c||}{conv5}			
&\multicolumn{1}{c|}{3$\times$3}
&\multicolumn{1}{c|}{2}
&\multicolumn{1}{c|}{96 / 128}
&\multicolumn{1}{c|}{conv4\_2} \\		
	
\multicolumn{1}{|c||}{conv6}			
&\multicolumn{1}{c|}{3$\times$3}
&\multicolumn{1}{c|}{2}
&\multicolumn{1}{c|}{128 / 192}
&\multicolumn{1}{c|}{conv5} \\
\hline
\end{tabular}}
\end{table*}

\begin{table*}[t]
\centering
\caption{The network details of the descriptor matching unit (M) of NetE in LiteFlowNet at pyramid level 5. ``upconv'', ``f-warp'', ``corr'', and ``loss'' denote the fractionally strided convolution (so-called deconvolution), feature warping, correlation, and the layer where training loss is applied, respectively. Furthermore, ``conv5a' and ``conv5b'' denote the high-dimensional features of images $I_{1}$ and $I_{2}$ generated from NetC at pyramid level 5.} \label{tab:NetE-M-L5}
\scalebox{1}{
\begin{tabular}{cccccc}
\hline

\multicolumn{1}{|c||}{Layer name} 			
&\multicolumn{1}{c|}{Kernel}
&\multicolumn{1}{c|}{Stride}									
&\multicolumn{1}{c|}{\# Ch. In / Out}
&\multicolumn{1}{c|}{Input(s)} \\
\hline\hline

\multicolumn{1}{|c||}{upconv5\_M}			
&\multicolumn{1}{c|}{4$\times$4}
&\multicolumn{1}{c|}{0.5}
&\multicolumn{1}{c|}{2 / 2} 
&\multicolumn{1}{c|}{flow6\_R} \\
																		
\multicolumn{1}{|c||}{f-warp5\_M}			
&\multicolumn{1}{c|}{-}
&\multicolumn{1}{c|}{-}
&\multicolumn{1}{c|}{(128, 2) / 128}
&\multicolumn{1}{c|}{conv5b,  upconv5\_M} \\

\multicolumn{1}{|c||}{corr5\_M}			
&\multicolumn{1}{c|}{1$\times$1}
&\multicolumn{1}{c|}{1}
&\multicolumn{1}{c|}{(128, 128) / 49}
&\multicolumn{1}{c|}{conv5a, f-warp5\_M} \\

\multicolumn{1}{|c||}{conv5\_1\_M}			
&\multicolumn{1}{c|}{3$\times$3}
&\multicolumn{1}{c|}{1}
&\multicolumn{1}{c|}{49 / 128}
&\multicolumn{1}{c|}{corr5\_M} \\

\multicolumn{1}{|c||}{conv5\_2\_M}			
&\multicolumn{1}{c|}{3$\times$3}
&\multicolumn{1}{c|}{1}
&\multicolumn{1}{c|}{128 / 64}
&\multicolumn{1}{c|}{conv5\_1\_M} \\

\multicolumn{1}{|c||}{conv5\_3\_M}			
&\multicolumn{1}{c|}{3$\times$3}
&\multicolumn{1}{c|}{1}
&\multicolumn{1}{c|}{64 / 32}
&\multicolumn{1}{c|}{conv5\_2\_M} \\	
	
\multicolumn{1}{|c||}{conv5\_4\_M}			
&\multicolumn{1}{c|}{3$\times$3}
&\multicolumn{1}{c|}{1}
&\multicolumn{1}{c|}{32 / 2}
&\multicolumn{1}{c|}{conv5\_3\_M} \\
	
\multicolumn{1}{|c||}{flow5\_M, loss5\_M}			
&\multicolumn{2}{c|}{element-wise sum}
&\multicolumn{1}{c|}{(2, 2) / 2}
&\multicolumn{1}{c|}{upconv5\_M, conv5\_4\_M} \\						 
\hline
\end{tabular}}
\end{table*}

\begin{table*}[t]
\centering
\caption{Network details of the sub-pixel refinement unit (S) of NetE in LiteFlowNet at pyramid level 5.} \label{tab:NetE-S-L5}
\scalebox{1}{
\begin{tabular}{ccccc}
\hline

\multicolumn{1}{|c||}{Layer name} 			
&\multicolumn{1}{c|}{Kernel}
&\multicolumn{1}{c|}{Stride}									
&\multicolumn{1}{c|}{\# Ch. In / Out}

&\multicolumn{1}{c|}{Input(s)} \\
\hline\hline
						
\multicolumn{1}{|c||}{f-warp5\_S}			
&\multicolumn{1}{c|}{-}
&\multicolumn{1}{c|}{-}
&\multicolumn{1}{c|}{(128, 2) / 128}
&\multicolumn{1}{c|}{conv5b, flow5\_M} \\

\multicolumn{1}{|c||}{conv5\_1\_S}			
&\multicolumn{1}{c|}{3$\times$3}
&\multicolumn{1}{c|}{1}
&\multicolumn{1}{c|}{258 / 128}
&\multicolumn{1}{c|}{conv5a, f-warp5\_S, flow5\_M} \\

\multicolumn{1}{|c||}{conv5\_2\_S}			
&\multicolumn{1}{c|}{3$\times$3}
&\multicolumn{1}{c|}{1}
&\multicolumn{1}{c|}{128 / 64}
&\multicolumn{1}{c|}{conv5\_1\_S} \\

\multicolumn{1}{|c||}{conv5\_3\_S}			
&\multicolumn{1}{c|}{3$\times$3}
&\multicolumn{1}{c|}{1}
&\multicolumn{1}{c|}{64 / 32}
&\multicolumn{1}{c|}{conv5\_2\_S} \\	
	
\multicolumn{1}{|c||}{conv5\_4\_S}			
&\multicolumn{1}{c|}{3$\times$3}
&\multicolumn{1}{c|}{1}
&\multicolumn{1}{c|}{32 / 2}
&\multicolumn{1}{c|}{conv5\_3\_S} \\
	
\multicolumn{1}{|c||}{flow5\_S, loss5\_S}			
&\multicolumn{2}{c|}{element-wise sum}
&\multicolumn{1}{c|}{(2, 2) / 2}
&\multicolumn{1}{c|}{flow5\_M, conv5\_4\_S} \\						 
\hline
\end{tabular}}
\end{table*}

\begin{table*}[t]
\centering
\caption{Network details of the flow regularization unit (R) of NetE in LiteFlowNet at pyramid level 5. ``rgb-warp'', ``norm'', ``negsq'', ``softmax'', and ``f-lcon'' denote the image warping, L2 norm of the RGB brightness difference between the two input images, negative-square, normalized exponential operation over each $1\times 1\times~$(\#~Ch.~In) column in the 3-D tensor, and feature-driven local convolution, respectively. Furthermore, ``conv\_dist'' that highlights the output of the convolution layer is used as the feature-driven distance metric $\mathcal{D}$ Eq.~(7) in the main manuscript. ``im5a'' and ``im5b'' denote the down-sized images of $I_{1}$ and $I_{2}$ at pyramid level 5, respectively} \label{tab:NetE-R-L5}
\scalebox{1}{
\begin{tabular}{ccccc}
\hline

\multicolumn{1}{|c||}{Layer name} 			
&\multicolumn{1}{c|}{Kernel}
&\multicolumn{1}{c|}{Stride}									
&\multicolumn{1}{c|}{\# Ch. In / Out}

&\multicolumn{1}{c|}{Input(s)} \\
\hline\hline
						
\multicolumn{1}{|c||}{rm-flow5\_R}			
&\multicolumn{2}{c|}{remove mean}
&\multicolumn{1}{c|}{2 / 2}
&\multicolumn{1}{c|}{flow5\_S} \\

\multicolumn{1}{|c||}{rgb-warp5\_R}			
&\multicolumn{1}{c|}{-}
&\multicolumn{1}{c|}{-}
&\multicolumn{1}{c|}{(3, 2) / 3}
&\multicolumn{1}{c|}{im5b, flow5\_S} \\

\multicolumn{1}{|c||}{norm5\_R}			
&\multicolumn{2}{c|}{L2 norm}
&\multicolumn{1}{c|}{(3, 3) / 1}
&\multicolumn{1}{c|}{im5a, rgb-warp5\_R} \\

\multicolumn{1}{|c||}{conv5\_1\_R}			
&\multicolumn{1}{c|}{3$\times$3}
&\multicolumn{1}{c|}{1}
&\multicolumn{1}{c|}{131 / 128}
&\multicolumn{1}{c|}{conv5a, rm-flow5\_R, norm5\_R} \\

\multicolumn{1}{|c||}{conv5\_2\_R}			
&\multicolumn{1}{c|}{3$\times$3}
&\multicolumn{1}{c|}{1}
&\multicolumn{1}{c|}{128 / 128}
&\multicolumn{1}{c|}{conv5\_1\_R} \\

\multicolumn{1}{|c||}{conv5\_3\_R}			
&\multicolumn{1}{c|}{3$\times$3}
&\multicolumn{1}{c|}{1}
&\multicolumn{1}{c|}{128 / 64}
&\multicolumn{1}{c|}{conv5\_2\_R} \\	
	
\multicolumn{1}{|c||}{conv5\_4\_R}			
&\multicolumn{1}{c|}{3$\times$3}
&\multicolumn{1}{c|}{1}
&\multicolumn{1}{c|}{64 / 64}
&\multicolumn{1}{c|}{conv5\_3\_R} \\	
	
\multicolumn{1}{|c||}{conv5\_5\_R}			
&\multicolumn{1}{c|}{3$\times$3}
&\multicolumn{1}{c|}{1}
&\multicolumn{1}{c|}{64 / 32}
&\multicolumn{1}{c|}{conv5\_4\_R} \\	
	
\multicolumn{1}{|c||}{conv5\_6\_R}			
&\multicolumn{1}{c|}{3$\times$3}
&\multicolumn{1}{c|}{1}
&\multicolumn{1}{c|}{32 / 32}
&\multicolumn{1}{c|}{conv5\_5\_R} \\	

\multicolumn{1}{|c||}{conv5\_dist\_R}			
&\multicolumn{1}{c|}{3$\times$3}
&\multicolumn{1}{c|}{1}
&\multicolumn{1}{c|}{32 / 9}
&\multicolumn{1}{c|}{conv5\_6\_R} \\

\multicolumn{1}{|c||}{negsq5\_R}			
&\multicolumn{2}{c|}{negative-square}
&\multicolumn{1}{c|}{9 / 9}
&\multicolumn{1}{c|}{conv5\_dist\_R} \\

\multicolumn{1}{|c||}{softmax5\_R}			
&\multicolumn{1}{c|}{1$\times$1$\times$9}
&\multicolumn{1}{c|}{1}
&\multicolumn{1}{c|}{9 / 9}
&\multicolumn{1}{c|}{negsq5\_R} \\

\multicolumn{1}{|c||}{f-lcon5\_R (Out: flow5\_R), loss5\_R}			
&\multicolumn{1}{c|}{3$\times$3}
&\multicolumn{1}{c|}{1}
&\multicolumn{1}{c|}{(9, 2) / 2}
&\multicolumn{1}{c|}{softmax5\_R, flow5\_S} \\					 
\hline
\end{tabular}}
\end{table*}
{\small
\bibliographystyle{ieee}
\bibliography{egbib}
}
\end{document}